\documentclass[11pt]{article}

\usepackage[final]{acl}

\usepackage{times}
\usepackage{latexsym}

\usepackage[T1]{fontenc}

\usepackage[utf8]{inputenc}

\usepackage{microtype}
\usepackage{booktabs}
\usepackage{multirow}

\usepackage{inconsolata}

\usepackage{graphicx}
\usepackage{booktabs}   
\usepackage{amssymb}    
\usepackage{pifont}     
\usepackage{multirow}

\newcommand{\cmark}{\ding{51}} 
\newcommand{\xmark}{\ding{55}} 

\title{Understanding Mental States to Guide Social Influence in Multi-Person Group Dialogue}

\author{
  Zhichao Liang \quad Satoshi Nakamura \\
  The Chinese University of Hong Kong, Shenzhen \\
  \texttt{zhichaoliang@link.cuhk.edu.cn, snakamura@cuhk.edu.cn}
}

\begin{document}
\maketitle
\begin{abstract}
Existing dynamic Theory of Mind (ToM) benchmarks mostly place language models in a passive role: the model reads a sequence of connected scenarios and reports what people believe, feel, intend, and do as these states change. In real social interaction, ToM is also used for action: a speaker plans what to say in order to shift another person’s mental-state trajectory toward a goal. We introduce SocialMindChange, a benchmark that moves from tracking minds to changing minds in social interaction. Each instance defines a social context with 4 characters and five connected scenes. The model plays one character and generates dialogue across the five scenes to reach the target while remaining consistent with the evolving states of all participants. SocialMindChange also includes selected higher-order states. Using a structured four-step framework, we construct 1,200 social contexts, covering 6000 scenarios and over 90,000 questions, each validated for realism and quality. Evaluations on ten state-of-the-art LLMs show that their average performance is 54.2\% below human performance. This gap suggests that current LLMs still struggle to maintain and change mental-state representations across long, linked interactions. \footnote{The data and code will be released after the paper is accepted.}

\end{abstract}

\section{Introduction}

Theory of Mind (ToM) is the capacity to reason about what other people
believe, feel, intend, and do \cite{premack1978does, turner1988theory}.
As large language models (LLMs) are used more often in social settings—
such as chat, tutoring, counseling-style support, negotiation, and team coordination—
ToM is no longer only a supporting skill.
Models must track how human mental states change over time,
and they must also decide what to say next in order to reach a goal.
\begin{figure}[t]
    \centering
    \includegraphics[width=\columnwidth]{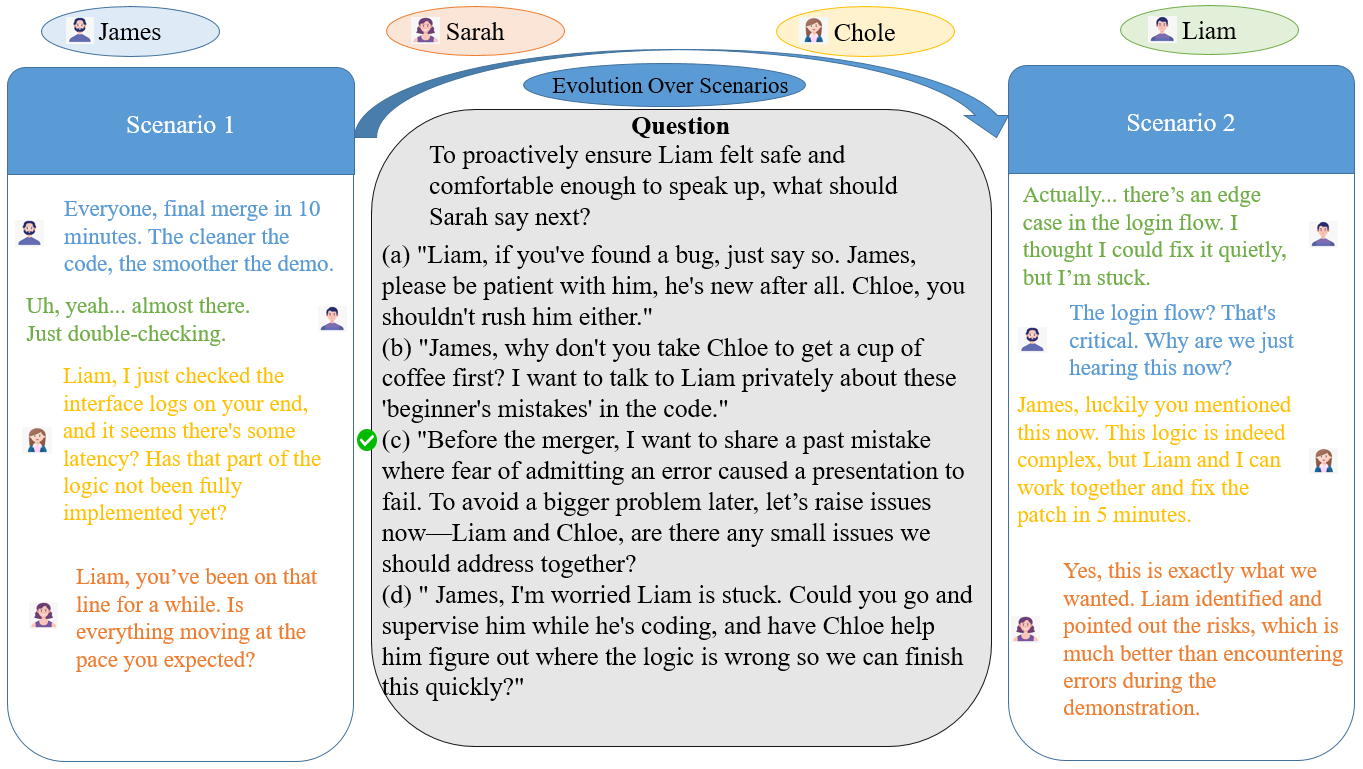}
    \caption{An simplified illustration of how proactive guidance operates in a four-agent setting.}
    \label{fig:example}
\end{figure}
Recent ToM benchmarks mainly test static snapshots:
a short story is given, and the model answers a question about one character’s mental state \cite{chen2024tombench, gandhi2023understanding, turner1988theory}.
The DYNTOM line of works move beyond this static setup by introducing
mental-state trajectories across temporally connected scenarios \cite{xiao2025towards}.
Instead of asking only what the belief, emotion, intention, or action is at a single point,
It also asks whether a state changes,
and how earlier states relate to later ones through several transformation question types.
This represents an important step toward testing ToM over connected social narratives.

However, DYNTOM still treats the model as an observer.
The model reads scenarios and answers questions,
but it does not need to act inside the interaction to influence how the trajectory evolves.
In real social settings, ToM is often used as a tool for
goal-directed social action:
a speaker chooses wording and timing to reduce anxiety,
build hope, calm conflict, or guide a group toward agreement.
Moreover, many real interactions are not purely dyadic.
When more people are involved,
mental states become interdependent:
what one person believes about another person’s belief
can be as important as the first-order state itself.
Such settings naturally require higher-order ToM
and longer-horizon planning.

To address these gaps, we introduce SocialMindChange,
a benchmark for strategic dynamic ToM in social, as shown in Figure~\ref{fig:example}.
SocialMindChange shifts the evaluation focus
from tracking mental-state trajectories to actively changing them.
Each instance defines a social context,
including the setting, character profiles, and their relationships,
and consists of a five-scene interaction.
The model must generate dialogue,
and when needed simple actions,
across scenes to move the focal person’s trajectory toward the target,
while remaining consistent with the evolving states of all characters.

SocialMindChange explicitly supports group reasoning.
Each scene includes four characters,
enabling realistic social dynamics such as competing intentions,
audience effects, and bystander reactions.
We annotate both first-order mental states
(belief, emotion, intention, and action for each character)
and selected higher-order states,
such as one character’s belief about another character’s intention. Evaluating ten representative LLMs—including the GPT-4 \cite{achiam2023gpt}, Llama 3 \cite{grattafiori2024llama}, Qwen 2 \cite{yang2025qwen3}, DeepSeek-V2 \cite{liu2024deepseek}, and GLM series \cite{glm2024chatglm}—we observe that their average performance is 54.2\% lower than human performance.
In summary, this paper makes three contributions:
\begin{itemize}
    \item A novel framework for evaluating LLMs’ dynamic Theory of Mind in social interaction, which tests whether LLMs can gradually moves a target person’s mental-state trajectory toward a specified outcome. We also provide a systematic data generation and validation process that constructs controlled social contexts and applies quality checks to ensure consistency and reliability of the evaluation data.
    \item A comprehensive benchmark with large-scale evaluations, designed to probe LLMs’ ability to plan supportive multi-turn dialogues that shift mental-state trajectories across a sequence of connected scenarios.
    \item Extensive empirical evaluation and failure analysis showing that current LLMs face clear difficulty in choosing supportive dialogue moves that satisfy competing group goals.
\end{itemize}

\section{Related Work}

\subsection{Theory of Mind Benchmarks for LLMs}

A large line of work studies whether large language models (LLMs) can perform
Theory of Mind (ToM) reasoning, often using classic false-belief tasks and
controlled question-answering settings \cite{nematzadeh2018evaluating, le2019revisiting, wu2023hi}.
Results consistently show that models still trail human performance,
and later analyses suggest that strong scores may rely on shortcuts rather than
stable ToM reasoning \cite{ullman2023large, shapira2024clever, kim2023fantom, sap2022neural}.
More recent benchmarks extend evaluation beyond belief to additional mental
state dimensions, including emotion, intention, and perception \cite{xu2024opentom, chen2024tombench, sabour2024emobench},
and some also incorporate multi-modal settings \cite{jin2024mmtom, shi2025muma, villa2025moments, bortoletto2025tom}.
However, most existing evaluations remain focused on single-scene snapshots
and do not directly test how mental states change across linked social
situations. SocialMindChange introduces a novel evaluation paradigm that diverges from existing research. Rather than merely describing mental states over time, our approach focuses on whether models can proactively select supportive dialogue strategies that align with evolving states and guide a desired social trajectory.

\subsection{Group Interaction and Higher-Order Theory of Mind}
Real-world social interactions often involve multiple participants,
where group dynamics, status concerns,
and indirect social signals influence outcomes.
In such settings, modeling only first-order mental states is often insufficient,
and higher-order Theory of Mind,
such as reasoning about what one person believes about another person’s belief,
becomes important \cite{dunbar200011, kim2023fantom, wu2023hi}.
Recent benchmarks show that nested mental-state reasoning remains challenging
for LLMs \cite{wu2023hi, gandhi2023understanding}, and related work explores ToM in multi-party settings \cite{zhou2023sotopia}. SocialMindChange brings these ideas together by modeling group interactions with
four characters per stage
and by including selected second- and third-order mental-state annotations.
This design tests whether models can maintain consistent nested mental states
over time while choosing supportive dialogue strategies that guide the target
trajectory without triggering new social conflict.

\subsection{Human Behavior Simulation and Data Construction with LLMs}

Large language models are increasingly used to generate realistic social
narratives and dialogue data in domains such as human--computer interaction \cite{hamalainen2023evaluating},
education \cite{wang2024towards}, and social science \cite{hua2023war, park2023generative, park2022social, aher2023using}.
Prior work shows that, with structured prompting,
LLMs can produce plausible human-like interaction traces.
At the same time, benchmark construction in this space requires careful design,
including consistency checks, human review,
and validation that questions test the intended capability rather than surface
patterns. SocialMindChange follows this general approach by using structured generation
to create controlled social contexts and temporally connected scenes,
combined with validation procedures that ensure dialogue content matches
annotated mental states and that multiple-choice options are realistic and
diagnostic.

\section{SocialMindChange}
\subsection{Building the SocialMindChange Benchmark: A Construction Framework}
\label{sec:construction}

\paragraph{Definitions and preliminaries.}
We first introduce the core concepts used in this work.
A Social Location is the physical setting where interaction takes  \cite{farrow2017social}.
It constrains behavior and shared norms during social exchange.
A Social Context defines the interaction setup and includes a social location,
four character profiles (such as basic demographics and personality traits),
and the relationships among them.
A Social Scenario is a single interaction snapshot at a specific time point.
In our benchmark, each social context contains five temporally linked scenarios,
which allows us to model how mental states evolve through ongoing interaction.
We refer to the complete unit---composed of the social location, social context,
and the full scenario sequence---as a Social Stage.

Each character’s first-order mental state is represented using four components:
belief, emotion, intention, and action.
When necessary to explain group dynamics, we also annotate selected
second- and third-order Theory of Mind (ToM) states,
such as ``A believes that B believes \ldots'' or
``A believes that B believes that C intends \ldots''.
These higher-order annotations are included only when they play a clear role
in shaping reactions or coordination among agents.

SocialMindChange evaluates proactive guidance rather than passive state recognition.
All evaluation tasks are framed as multiple-choice questions.
We ask what a speaker should say next,
what immediate state change is expected,
why the guidance works,
and which multi-step guidance plan best explains a target end state.

\begin{figure}[t]
    \centering
    \includegraphics[width=\columnwidth]{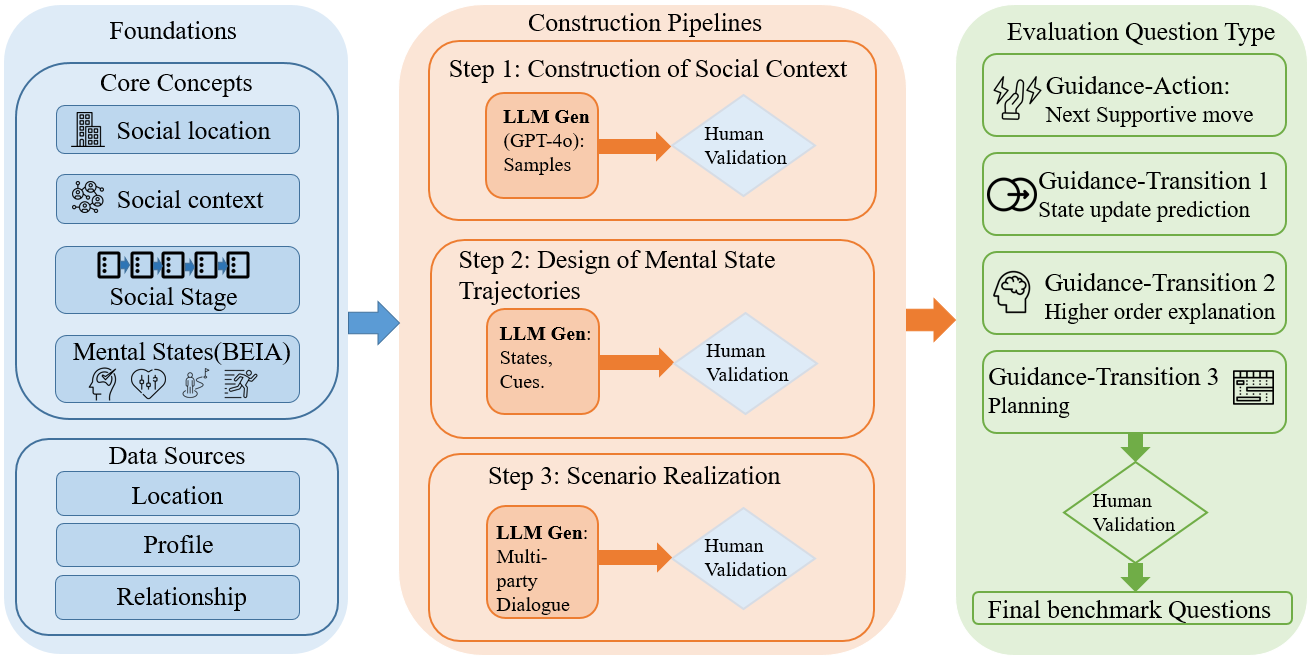}
    \caption{Benchmark Construction Framework \& Evaluation Question Generation}
    \label{fig:example2}
\end{figure}

\paragraph{Construction pipeline.}
Our benchmark is constructed through a three-step pipeline integrated with an Evaluation Question Generation module, as illustrated in Figure~\ref{fig:example2}.

\paragraph{Step 1: Construction of Social Context}
Each social context is composed of three elements: a social location, profiles for 4 characters, and the relationship structure among them. For social locations, following \cite{ziems2023normbank}, we collect 261 locations spanning 13 categories that cover common physical settings for everyday social interaction. 

For character profiles, we construct seven attribute pools—names, surnames, gender, occupation, education, race, and personality traits—using demographic statistics from the U.S. Census Bureau to better match realistic population distributions. For each social context, we randomly sample one location and generate 4 characters by drawing attributes independently from each pool.

Relationship structures are generated in two stages. We first design a small set of relationship exemplars manually, then prompt an LLM (GPT-4o) to produce a coherent multi-person relationship graph conditioned on the sampled profiles. The generated structure includes both pairwise relations (such as classmates, friends, or rivals) and group-level roles aligned with our task setting (for example, a target individual, a supportive guide, a competitive peer, and a supportive peer). 

To ensure quality, Two human annotators independently review each sampled profile and generated relationship structure. Any instance that is judged by at least one annotator to be unrealistic or inconsistent with the location or assigned roles is discarded. This validation process retains 89\% of the generated social contexts.

\paragraph{Step 2: Design of Mental State Trajectories}
For each validated social context, we design a sequence of five connected scenarios
in which the mental states of all characters evolve and influence one another.
The trajectory design follows a simple dependency structure from social psychology: 1) beliefs affect emotions;
2) beliefs and emotions shape intentions;
3) beliefs, emotions, and intentions jointly determine actions.

Using exemplar trajectories and explicit rules, we prompt an LLM to generate:
1) first-order state labels (belief, emotion, intention, action) for each
    character in each scenario;
2) selected second- and third-order ToM states that explain perception,
    misunderstanding, or group tension;
3) explicit transition cues between consecutive scenarios that describe
    what causes each state change (e.g., criticism, encouragement, public feedback,
    or reframing).

Trajectories are discarded if they miss state labels, lack transition cues,
or violate basic consistency constraints, such as abrupt belief changes without
explanation.
Human annotators then rate each remaining trajectory for coherence,
rationality, and authenticity (a 5-point scale).
Trajectories with mean scores below 4 are excluded (81.7\% retention rate).

\paragraph{Step 3: Scenario Realization with Multi-party Dialogue and Labels}
Based on the validated trajectories, each scenario is realized as
(i) a brief background description and
(ii) a multi-party dialogue involving all characters.
Dialogue is used as the primary format because it naturally exposes mental states
and mirrors real interaction. We prompt an LLM to generate dialogue under strict constraints:
1) utterances and actions must reflect the annotated belief, emotion,
    intention, and action;
2) group alignment or tension must be visible when higher-order states
    are involved;
3) the five scenarios must form a coherent narrative with clear temporal links.

Human annotators then evaluate each scenario for consistency with the designed
trajectory, coherence across the five scenarios, and authenticity of the interaction.
Scenarios that do not meet the 4.0 are removed (88.3\% are retained).

\subsection{Evaluation Question Types}
\label{sec:evaluation_questions}

Based on the validated scenarios and annotated mental-state trajectories,
we design evaluation questions to test whether a large language model can
provide proactive supportive guidance in temporally connected social interaction.
The question types are organized in a progressive manner,
moving from local action selection to long-horizon planning.
SocialMindChange includes four multiple-choice question types,
each targeting a different reasoning requirement
across the five-scenario interaction.

\paragraph{Guidance--Action (supportive next move under group constraints).}
Guidance--Action questions test whether a model can choose
the most appropriate next supportive utterance
for the guidance role at a given stage.
Each question specifies the current social context
together with explicit constraints,
such as preventing the target from disengaging
while maintaining participation from a competitive peer.
Answer options present plausible dialogue moves
that differ in how they balance
(i) the target’s dignity and agency,
(ii) group norms such as fairness and efficiency, and
(iii) the risk of social tension or escalation.

\paragraph{Guidance--Transition-1 (immediate group state update).}
Guidance--Transition-1 questions evaluate whether the model can predict
the immediate mental-state changes
that follow a specific supportive action.
Given an action selected at stage $t$,
the question asks for the most likely updates
from stage $t$ to stage $t{+}1$
across multiple characters,
including the target and other group members.
\paragraph{Guidance--Transition-2 (higher-order causal explanation).}
Guidance--Transition-2 questions assess whether the model understands
why a supportive move leads to a particular outcome,
using explicit mental-state mechanisms
and higher-order Theory of Mind.
\paragraph{Guidance--Transition-3 (full-trajectory guidance planning).}
Guidance--Transition-3 questions test long-horizon reasoning.
The model must select the best five-stage guidance plan
that leads to a desired end state,
such as the target becoming more hopeful and agentic
while group cohesion is maintained. The detailed templates are provided in Appendix~\ref{app:templates-a5}.

\paragraph{Options and ground truth.}
Answer options and ground-truth labels
are derived directly from the constructed
mental-state trajectories
and explicit transition cues.
Distractors are systematically generated from
alternative annotated states,
competing state values,
and plausible but incompatible causal links.
This design ensures that questions remain challenging
while still admitting a single unambiguous correct answer.
In addition, distractors are designed to fail in specific ways,
such as violating group constraints,
contradicting second- or third-order beliefs,
predicting incorrect multi-person state changes,
or undermining the intended long-term guidance objective.

\paragraph{Validation and evaluation.}
Each question is reviewed by human annotators
for clarity and answerability (5-point scale).
Items that do not meet 4.0
are regenerated.
For evaluation,
we report accuracy across all questions,
counting a prediction as correct
only if it exactly matches the ground-truth option. We calculate the percentage of correct answers.

\begin{table}[t]
\centering
\small
\setlength{\tabcolsep}{3pt}
\begin{tabular}{lccccccccr}
\toprule
\textbf{Benchmark} &
\textbf{P} &
\textbf{L} &
\textbf{Pr} &
\textbf{R} &
\textbf{D} &
\textbf{H} &
\textbf{M} &
\textbf{Pro} &
\textbf{\#Q} \\
\midrule
ToMi        & \xmark & \xmark & \xmark & \xmark & \xmark & \cmark & \cmark & \xmark & 999 \\
SocialIQA   & \xmark & \xmark & \xmark & \xmark & \xmark & \xmark & \xmark & \xmark & 37{,}588 \\
Hi-ToM      & \xmark & \cmark & \xmark & \xmark & \xmark & \cmark & \cmark & \xmark & 1{,}200 \\
OpenToM     & \xmark & \xmark & \cmark & \xmark & \xmark & \cmark & \xmark & \xmark & 2{,}384 \\
BigToM      & \cmark & \cmark & \xmark & \xmark & \cmark & \xmark & \xmark & \xmark & 600 \\
TOMBENCH    & \cmark & \xmark & \xmark & \xmark & \xmark & \cmark & \cmark & \xmark & 2{,}860 \\
DYNTOM    & \cmark & \cmark & \cmark & \cmark & \cmark & \xmark & \xmark & \xmark & 78{,}100 \\
\midrule
\textbf{SocialMindChange} &
\cmark & \cmark & \cmark & \cmark & \cmark & \cmark & \cmark & \cmark & \textbf{96{,}000} \\
\bottomrule
\end{tabular}
\caption{Comparison with Existing Benchmarks. (P: plot; L: location; Pr: profile; R: relationship; D: dynamic states;
H: higher-order ToM; M: multi-party interaction;
Pro: proactive guidance; \#Q: number of questions)}
\label{tab:benchmark_comparison_extended}
\end{table}

\subsection{Statistics}
The final SocialMindChange benchmark contains 1,200 validated social stages. Each stage is composed of a social location (physical setting), a social context with 4 characters and their relationships, and a sequence of five temporally connected social scenarios.

For evaluation, we construct a multiple-choice question set for every social stage. Unlike benchmarks that focus only on mental-state reporting, SocialMindChange is designed to assess proactive supportive guidance and its effects on both first-order and selected higher-order mental states. Specifically, for each social stage we generate 80 questions spanning four question types (Guidance–Action and Guidance–Transition-1/2/3), resulting in 96,000 questions overall. Our benchmark, SocialMindChange, exhibits several distinct advantages over previous ToM evaluations (see Table~\ref{tab:benchmark_comparison_extended}).

\section{Experiments}

\subsection{Experimental Setup}
\label{sec:experimental_setup}

\paragraph{Models.}
We evaluate SocialMindChange on nine large language models covering a wide range of scales (7B–70B) and architectures. These include both closed- and open-source models: GPT-4o, Llama-3.1 (8B, 70B), Mistral-7B, Mixtral-8x7B, Qwen2 (7B, 72B), DeepSeek-V2, and GLM-4. Models are accessed via official APIs or public weights.

\paragraph{Prompting settings.}
We consider two settings: vanilla prompting, where models directly choose an answer, and zero-shot chain-of-thought prompting, where models reason step by step before answering~\citep{wei2022chain}. For both settings, we use temperature = 0.7 and top-$p$ = 0.9 for all models to keep the comparison consistent. Prompting templates appear in Appendix~\ref{app:prompt-templates-b1}.

\paragraph{Human baseline.}
Human performance is estimated using Five graduate students not involved in data creation. They answer a random 20\% subset of the benchmark. We report a human baseline as the average over five annotators, with standard deviations reflecting inter-annotator agreement.

\subsection{Experimental Results}
Table~\ref{tab:main_results} presents the performance of large language models
on SocialMindChange across the four evaluation question types,
namely Guidance--Action and Guidance--Transition-1/2/3,
under both vanilla prompting and chain-of-thought (CoT) prompting.
Human performance is reported as the average accuracy
across Five annotators.
Below, we highlight several key observations
drawn from the experimental results.

\begin{table*}[t]
\centering
\small
\setlength{\tabcolsep}{3pt}
\begin{tabular}{lcc cc cc cc c}
\toprule
\multirow{2}{*}{\textbf{Subject}} &
\multicolumn{2}{c}{\textbf{Belief}} &
\multicolumn{2}{c}{\textbf{Emotion}} &
\multicolumn{2}{c}{\textbf{Intention}} &
\multicolumn{2}{c}{\textbf{Action}} &
\multirow{2}{*}{\textbf{AVG.}} \\
& \textbf{GA} & \textbf{GT}
& \textbf{GA} & \textbf{GT}
& \textbf{GA} & \textbf{GT}
& \textbf{GA} & \textbf{GT} & \\
\midrule
Human
& $81.3_{\pm10.1}$ & $79.2_{\pm9.4}$
& $85.5_{\pm12.4}$ & $77.7_{\pm10.3}$
& $80.4_{\pm17.2}$ & $75.1_{\pm16.6}$
& $72.5_{\pm13.1}$ & $75.3_{\pm11.4}$
& $76.6_{\pm10.9}$ \\
\midrule
GPT-4o
& \textbf{62.4} & \textbf{33.1}
& \textbf{73.5} & \textbf{34.2}
& \textbf{68.9} & \textbf{38.5}
& \textbf{76.8} & \textbf{41.2}
& \textbf{48.9} \\
Llama-3.1-70B
& 45.6 & 34.5
& 73.2 & 32.8
& 64.3 & 32.9
& 73.5 & 33.1
& 36.9 \\
Llama-3.1-8B
& 26.4 & 14.8
& 35.2 & 16.5
& 18.9 & 13.8
& 22.1 & 12.6
& 17.0 \\
Mixtral-8x7B
& 19.5 & 18.2
& 39.8 & 15.1
& 23.4 & 8.5
& 30.6 & 7.8
& 15.2 \\
Mistral-7B
& 13.1 & 9.4
& 19.2 & 8.5
& 13.5 & 6.7
& 15.1 & 7.5
& 9.9 \\
Qwen2-72B
& 53.8 & 28.5
& 70.2 & 32.4
& 64.1 & 25.8
& 77.5 & 15.2
& 33.4 \\
Qwen2-7B
& 18.8 & 17.5
& 37.5 & 17.1
& 15.6 & 13.9
& 20.8 & 12.4
& 16.6 \\
DeepSeek-V2
& 5.1 & 8.8
& 3.9 & 3.8
& 2.9 & 6.1
& 2.1 & 4.5
& 4.2 \\
GLM-4
& 24.8 & 16.5
& 38.1 & 17.2
& 18.3 & 10.8
& 35.6 & 13.9
& 19.2 \\
\midrule
\textbf{LLM AVG.}
& 29.9 & 20.1
& 35.6 & 19.7
& 32.2 & 17.4
& 36.3 & 16.5
& 22.4 \\
\midrule
GPT-4o+CoT
& 59.1 & \textbf{33.8}
& 70.5 & \textbf{36.2}
& \textbf{68.5} & \textbf{35.4}
& 72.8 & \textbf{40.1}
& \textbf{47.5} \\
Llama-3.1-70B+CoT
& 50.8 & 29.5
& \textbf{75.4} & 33.1
& 64.5 & 32.4
& \textbf{75.2} & 35.8
& 44.8 \\
Llama-3.1-8B+CoT
& 27.2 & 13.1
& 38.5 & 17.4
& 18.5 & 13.2
& 15.8 & 13.1
& 16.9 \\
Mixtral-8x7B+CoT
& 15.8 & 11.5
& 25.2 & 11.2
& 21.9 & 7.2
& 21.5 & 5.1
& 10.7 \\
Mistral-7B+CoT
& 18.2 & 8.4
& 15.1 & 9.1
& 16.5 & 7.8
& 15.9 & 4.2
& 8.4 \\
Qwen2-72B+CoT
& \textbf{61.8} & 32.8
& 68.4 & 31.5
& 65.2 & 26.4
& 76.1 & 21.5
& 36.8 \\
Qwen2-7B+CoT
& 24.2 & 15.2
& 38.1 & 17.1
& 25.4 & 16.5
& 14.1 & 15.2
& 18.1 \\
DeepSeek-V2+CoT
& 6.1 & 9.2
& 2.6 & 8.8
& 4.1 & 6.7
& 4.2 & 5.0
& 5.3 \\
GLM-4+CoT
& 27.5 & 21.8
& 41.5 & 18.4
& 29.8 & 14.5
& 37.1 & 13.9
& 20.8 \\
\midrule
\textbf{LLM+CoT AVG.}
& 32.3 & 19.5
& 41.7 & 20.3
& 34.9 & 17.8
& 37.0 & 17.1
& 23.3 \\
\bottomrule
\end{tabular}
\caption{Performance on SocialMindChange for both LLMs and humans.
GA denotes Guidance–Action and GT denotes Guidance–Transition.
All values are accuracy (\%).}
\label{tab:main_results}
\end{table*}

\paragraph{Comparison Between Humans and LLMs}
Human annotators achieve an average accuracy of 76.6\% across all tasks.
In contrast, all evaluated large language models perform far below this baseline:
the LLM average accuracy is 22.4\%,
which is 54.2 percentage points lower than human performance.
Even the strongest model under vanilla prompting, GPT-4o (48.9\%),
still trails humans by a substantial margin
(76.6\% vs.\ 48.9\%, a gap of 27.7 points).
The performance gap is especially pronounced for
Guidance--Transition (GT) questions.
On average, LLMs achieve 29.9--36.3\% accuracy on
Guidance--Action (GA) questions across the four mental-state targets,
but only 16.5--20.1\% on GT questions,
indicating a sharp drop when the task requires predicting or explaining
how mental states change across stages.
A similar pattern is observed in the strongest model:
GPT-4o reaches 62.4--76.8\% accuracy on GA,
but only 33.1--41.2\% on GT.
Chain-of-thought prompting leads to only a modest overall change
in average performance
(22.4\% $\rightarrow$ 23.3\%),
although it benefits some individual models,
such as Llama-3.1-70B (36.9\% $\rightarrow$ 44.8\%)
and Qwen2-72B (33.4\% $\rightarrow$ 36.8\%).

Finally, while LLMs can sometimes approach or even exceed
human performance on local GA subsets
(e.g., GPT-4o on Action--GA: 76.8\% vs.\ human 72.5\%),
this advantage does not carry over to GT questions.
This result suggests that selecting a reasonable next move
is substantially easier than maintaining
temporally consistent, multi-person mental-state updates.

\paragraph{Performance Differences Among LLMs in ToM Tasks}

Under vanilla prompting, GPT-4o achieves the highest accuracy on SocialMindChange (48.9\%), followed by Llama-3.1-70B (36.9\%). Among open-source models, Llama-3.1-70B and Qwen2-72B perform the best, yet all evaluated models remain far below the human baseline (76.6\%). Across systems, performance is consistently much higher on GA than on GT, indicating that models can select locally reasonable responses but struggle with proactive guidance that shapes future mental-state trajectories. This difficulty is especially pronounced in multi-person settings and when higher-order Theory of Mind is required. Overall, the results suggest that current large language models lack a reliable ability to actively guide group dynamics over multiple steps, and instead mainly react to the immediate context.

\paragraph{GA vs. GT}
Table~\ref{tab:main_results} reveals a consistent gap between GA and GT. Across models, GA accuracy ranges from 29.9–36.3\%, while GT drops to 16.5–20.1\%, a gap of about 13–16 points, with the largest decline in emotion (36.3\% → 20.1\%). This reflects the different demands of the two tasks: GA tests selection of a locally appropriate supportive move, whereas GT requires proactive guidance that anticipates multi-step, multi-person mental-state changes, including higher-order beliefs. The sharp drop shows that current LLMs can react reasonably in the moment but struggle to guide group mental-state trajectories over time.

\begin{figure}[t]
    \centering
    \includegraphics[width=\columnwidth]{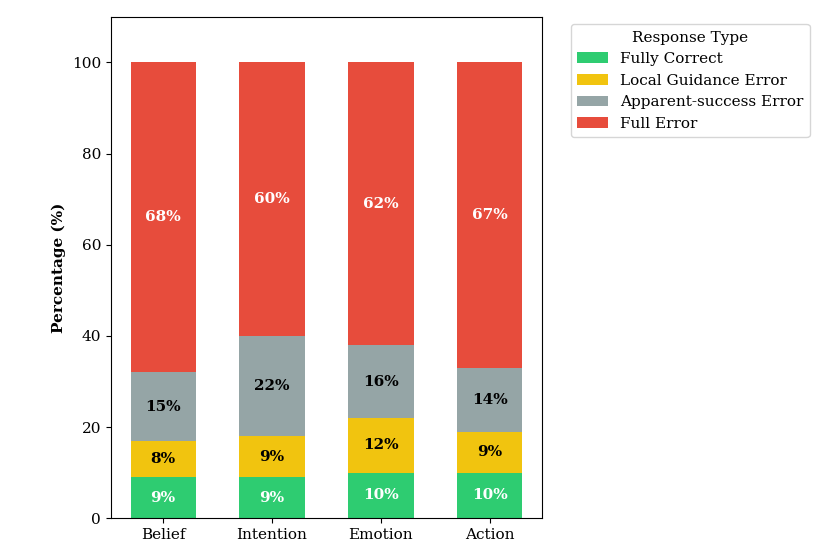}
    \caption{Performance Breakdown of GPT-4o on ToM Tasks.}
    \label{fig:example3}
\end{figure}

\paragraph{Effect of Chain-of-Thought Prompting}

Chain-of-thought (CoT) prompting has mixed and model-dependent effects on SocialMindChange, with only a small average gain (+0.9). While CoT helps some models (e.g., Llama-3.1-70B and Qwen2-72B), it hurts others, including GPT-4o and Mixtral-8×7B.  CoT can support local reasoning about what to say next, but the task requires anticipating how an intervention will shape future mental-state trajectories of multiple people, including higher-order beliefs. Generic CoT often treats each stage separately, which weakens reasoning about how a guiding action alters later group dynamics. As a result, CoT does not consistently improve—and can even degrade—performance when proactive, multi-step mental-state guiding is required, pointing to the need for prompts that explicitly enforce cross-stage and group-consistent guidance reasoning.

\paragraph{Mental-State-Specific Performance Differences}
Table~\ref{tab:main_results} shows clear differences in models’ performance across mental-state targets, and the pattern is most visible for GA. Averaged across LLMs under vanilla prompting, action-focused GA achieves the highest accuracy (36.3\%), followed closely by emotion (35.6\%) and intention (32.2\%), while belief-focused GA is the lowest (29.9\%). This yields a 6.4-point gap between action and belief (36.3\% vs. 29.9\%).

We hypothesize that these differences reflect how much each target can be recovered from the dialogue and constraints. In GA, selecting a supportive next move for action can often be guided by explicit norms and immediate interaction goals (e.g., reduce shame, avoid escalation, keep the group engaged), whereas belief is more often implicit and requires inference from indirect evidence and social interpretation.

\subsection{Further Analysis}
\paragraph{LLMs’ Limits in Proactive Supportive Guidance for GT}

Reasoning about mental states across temporally linked stages
requires models to go beyond identifying current beliefs or emotions
and to predict how proactive supportive guidance
reshapes future group dynamics.
In SocialMindChange,
this challenge is most visible in
GT tasks (Transition-1/2/3),
where models must infer cross-stage effects,
including how the target’s belief, emotion, intention, and action change,
how peers respond,
and how higher-order beliefs are updated.

To analyze failure modes,
we organize related questions into
\emph{guidance-dependency sets}.
A primary transition question ($C$),
which asks how mental states change
from stage $t$ to stage $t{+}1$ as a direct consequence of a chosen supportive move.,
depends on a set of prerequisite questions ($D$)
that test:
(i) correct grounding of mental states at stage $t$;
(ii) understanding the social meaning of the supportive move; and
(iii) tracking higher-order beliefs
that shape outcomes such as reduced shame or increased tension.

Based on this dependency structure,
we classify model behavior into four categories:
\emph{fully correct},
\emph{local guidance error}
(correct grounding but incorrect transition),
\emph{apparent-success error}
(correct transition despite grounding errors), and
\emph{full error}.

Analysis of GPT-4o results (Figure~\ref{fig:example3})
shows that fully correct cases are rare,
accounting for only 9--10\% of instances.
Full errors dominate,
covering 60--68\% of cases,
especially for belief and action,
indicating difficulty with both state identification
and guidance-driven change.
Local guidance errors (8--12\%)
demonstrate that even when mental-state grounding is correct,
models often fail to predict
how supportive guidance reshapes the next stage.
Apparent-success errors (14--22\%)
further suggest that some correct answers
may arise from shallow patterns
rather than stable reasoning.

Overall,
current large language models struggle
to integrate mental-state grounding,
the social meaning of guidance,
and higher-order belief updates
into a coherent cross-stage prediction.
These results reveal clear limits
in proactive supportive guidance
for temporally extended, multi-person social interaction.

\begin{figure}[t]
    \centering
    \includegraphics[width=\columnwidth]{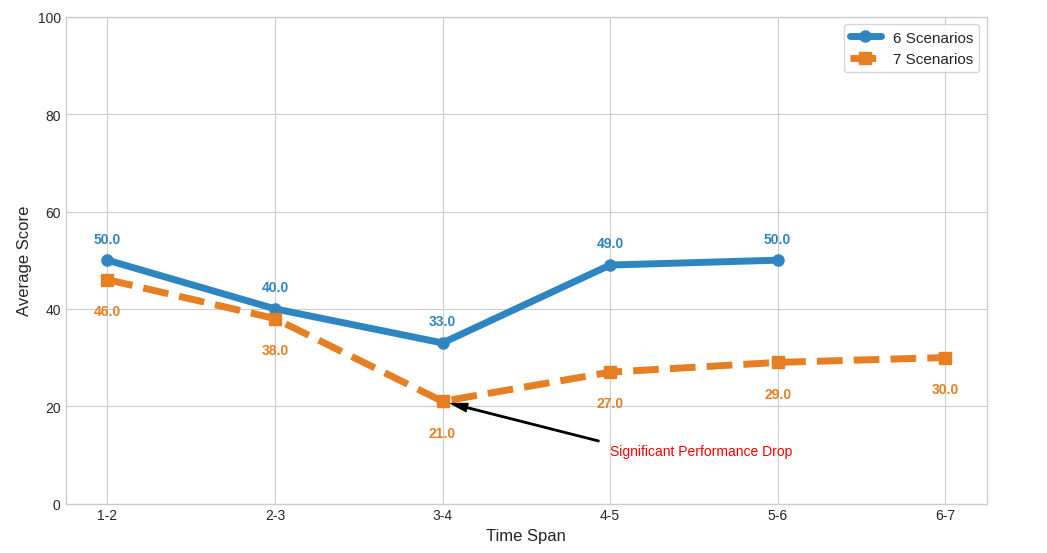}
    \caption{GPT-4o average score (scenarios 6–7)}
    \label{fig:example4}
\end{figure}
\begin{table}[t]
\centering
\small
\setlength{\tabcolsep}{7pt}
\begin{tabular}{lcccc}
\toprule
\textbf{Category} & \textbf{Time Span} &
\textbf{w/o T} &
\textbf{w/ T} &
${\Delta}$ \\
\midrule
\multirow{3}{*}{5 scenarios}
& \textbf{1--2} & 47.0 & 48.0 & +1.0 \\
& \textbf{2--3} & 46.0 & 48.0 & +2.0 \\
& \textbf{3--4} & 46.0 & 47.0 & +1.0 \\
\midrule
\multirow{3}{*}{6 scenarios}
& \textbf{1--2} & 50.0 & 52.0 & +2.0 \\
& \textbf{2--3} & 40.0 & 45.0 & +5.0 \\
& \textbf{3--4} & 33.0 & 41.0 & +8.0 \\
\midrule
\multirow{3}{*}{7 scenarios}
& \textbf{1--2} & 46.0 & 46.0 & +0.0 \\
& \textbf{2--3} & 38.0 & 45.0 & +7.0 \\
& \textbf{3--4} & 21.0 & 35.0 & +14.0 \\
\bottomrule
\end{tabular}
\caption{GPT-4o performance comparison with and without truncation, evaluated across different time spans and categories (5, 6, and 7 total scenarios).
$\Delta$ indicates the performance difference between the two settings.}
\label{tab:truncation_analysis}
\end{table}

\paragraph{LLMs’ Difficulty in Middle-Stage Reasoning}
GT questions are consistently harder than GA questions, showing that models struggle to predict how supportive guidance affects mental states over time. To pinpoint this issue, we analyze performance \emph{time span}, defined as the transition between consecutive scenarios.

Figure~\ref{fig:example4} reveals a clear U-shaped pattern
that becomes more pronounced as sequence length increases.
For six-scenario sequences,
accuracy drops from 50.0\% at span 1--2
to 40.0\% at span 2--3,
reaches a minimum of 33.0\% at span 3--4,
and then recovers to 49.0\% at span 4--5
and 50.0\% at span 5--6.
The pattern is more severe for seven-scenario sequences:
performance decreases from 46.0\% at span 1--2
and 38.0\% at span 2--3
to 21.0\% at span 3--4,
followed by only partial recovery
(27.0\% at span 4--5,
29.0\% at span 5--6,
and 30.0\% at span 6--7).
These results indicate that failures
concentrate at middle-stage transitions.
To test the ``lost in the middle'' explanation,
we conduct a truncation experiment
that retains only the first four scenarios at inference time.
As reported in Table~\ref{tab:truncation_analysis},
truncation yields the largest gains at the middle spans.
For seven-scenario sequences,
accuracy improves by 14.0 points on span 3--4
(21.0\% $\rightarrow$ 35.0\%)
and by 7.0 points on span 2--3
(38.0\% $\rightarrow$ 45.0\%).
For six-scenario sequences,
truncation also improves span 3--4
(33.0\% $\rightarrow$ 41.0\%, +8.0)
and span 2--3
(40.0\% $\rightarrow$ 45.0\%, +5.0).
In contrast, gains for five-scenario sequences
are small, ranging from +1.0 to +2.0 points.

Overall, these results expose a clear limitation on SocialMindChange: models handle early transitions and sometimes late ones, but fail to maintain coherent mental-state updates through the middle of longer interactions, where proactive supportive guidance must manage rising tension, group reactions, and higher-order beliefs.

\begin{table}[t]
\centering
\small
\setlength{\tabcolsep}{10pt}
\begin{tabular}{lccc}
\toprule
\textbf{Model} & \textbf{4-Person} & \textbf{5-Person} & \textbf{6-Person} \\
\midrule
GPT-4o & 48.9 & 45.3 & 37.5 \\
\bottomrule
\end{tabular}
\caption{Accuracy (\%) of GPT-4o across varying group sizes (4--6).}
\label{tab:size_scaling_overall}
\end{table}

\paragraph{Scaling to Larger Groups}
SocialMindChange currently uses four participants per stage, which already requires the model to track multi-person first-order states and selected higher-order beliefs. To evaluate whether current LLMs can generalize proactive supportive guidance to more realistic group settings, we further extend the benchmark to larger groups with 5–6 participants per stage. Concretely, we keep the same five-scenario temporal structure, but add one or two additional bystanders with distinct interaction roles. The experimental results are shown in Table~\ref{tab:size_scaling_overall}, and it can be seen that performance decreases as the number of participants  (48.9\% → 45.3\% → 37.5\%).

\section{Conclusion}
We introduce SocialMindChange, a benchmark for evaluating LLMs’ Theory of Mind in dynamic social interactions with a focus on proactive supportive guidance. The benchmark tests whether models can reason about how dialogue moves change multi-person mental-state trajectories over time. While human performance reaches 76.6\%, the best LLM achieves only 48.9\%, leaving a 27.7-point gap, most pronounced on Guidance–Transition questions that require cross-stage reasoning. SocialMindChange provides a clear framework for measuring and improving LLMs’ ability to guide evolving mental states in realistic social settings.

\section*{Limitations}
\paragraph{Limited Prompting and Inference Strategies.}
We primarily compare vanilla prompting and standard chain-of-thought prompting under a unified output format for automatic evaluation. However, proactive supportive guidance may benefit from prompting and decoding strategies that explicitly track cross-stage dependencies and multi-agent constraints. Future work could evaluate alternative methods such as think-twice style re-evaluation \cite{wilf2024think}, self-consistency\cite{wang2022self}, and other structured prompting approaches that encourage models to reason about how a chosen supportive move changes downstream mental states and group reactions.
\paragraph{Limited Model Coverage.}
SocialMindChange is designed to evaluate proactive supportive guidance in multi-person social interactions, but our empirical study is constrained by compute and budget. We therefore evaluate a set of nine widely used LLMs that are representative of both closed-source and open-weight families. While this coverage supports controlled comparisons, it does not exhaust the rapidly evolving model landscape. In particular, claude-series models and emerging open-source releases may exhibit different strengths in long-horizon guidance planning and higher-order reasoning. Expanding model coverage is an important direction for future work.
\paragraph{Limited State Scope and Modalities.}
Our benchmark operationalizes mental-state dynamics using four core fields---belief, emotion, intention, and action---and includes selected second- and third-order beliefs for a subset of cases. Although this design captures key psychological dependencies and supports scalable annotation, it does not cover the full range of social-cognitive variables. Future work could extend the state space to include additional constructs. In addition, SocialMindChange currently focuses on text-based interactions. Real-world supportive guidance also relies on nonverbal signals (e.g., facial expressions, tone, timing), so extending the benchmark to multimodal settings with visual and auditory cues would provide a more complete test of models’ ability to track and shape mental-state trajectories over time.

\section*{Ethical considerations}
Our data validation framework was designed with a human-centric approach, ensuring that annotator well-being was prioritized in accordance with standard ethical practices. Annotators were informed that they could pause or stop the annotation process at any time if they encountered content that caused discomfort. We compensated annotators at a rate of approximately 3× their local hourly minimum wage. To reduce potential harms, we instructed annotators to check the benchmark data for realism and consistency while keeping the content free from unsafe, toxic, biased, offensive, or otherwise harmful material. We also encourage careful handling of sensitive social situations (e.g., conflict, shame, exclusion) by focusing on everyday, non-graphic scenarios and by avoiding targeted hate or harassment. All language models used in experiments were accessed and used in accordance with their official usage policies and intended purposes.

Theory of Mind is a human social-cognitive capacity, and benchmarking LLMs on SocialMindChange may invite anthropomorphic interpretations (e.g., treating a model as if it “understands” minds in the human sense). We emphasize that SocialMindChange is not intended to claim that LLMs possess human mental states or human-like consciousness. Instead, our goal is to measure whether models can produce behavior that is consistent with mental-state reasoning in multi-stage social interactions—especially when providing proactive supportive guidance and anticipating how such guidance changes beliefs, emotions, intentions, and actions over time. We view this as a practical evaluation target for safer and more reliable human–AI interaction, not as evidence that models have genuine human Theory of Mind.



\bibliography{acl_latex}

@article{premack1978does,
  title={Does the chimpanzee have a theory of mind?},
  author={Premack, David and Woodruff, Guy},
  journal={Behavioral and brain sciences},
  volume={1},
  number={4},
  pages={515--526},
  year={1978},
  publisher={Cambridge University Press}
}

@book{turner1988theory,
  title={A theory of social interaction},
  author={Turner, Jonathan H},
  year={1988},
  publisher={Stanford University Press}
}

@article{chen2024tombench,
  title={Tombench: Benchmarking theory of mind in large language models},
  author={Chen, Zhuang and Wu, Jincenzi and Zhou, Jinfeng and Wen, Bosi and Bi, Guanqun and Jiang, Gongyao and Cao, Yaru and Hu, Mengting and Lai, Yunghwei and Xiong, Zexuan and others},
  journal={arXiv preprint arXiv:2402.15052},
  year={2024}
}

@article{gandhi2023understanding,
  title={Understanding social reasoning in language models with language models},
  author={Gandhi, Kanishk and Fr{\"a}nken, Jan-Philipp and Gerstenberg, Tobias and Goodman, Noah},
  journal={Advances in Neural Information Processing Systems},
  volume={36},
  pages={13518--13529},
  year={2023}
}

@article{xiao2025towards,
  title={Towards Dynamic Theory of Mind: Evaluating LLM Adaptation to Temporal Evolution of Human States},
  author={Xiao, Yang and Wang, Jiashuo and Xu, Qiancheng and Song, Changhe and Xu, Chunpu and Cheng, Yi and Li, Wenjie and Liu, Pengfei},
  journal={arXiv preprint arXiv:2505.17663},
  year={2025}
}

@article{achiam2023gpt,
  title={Gpt-4 technical report},
  author={Achiam, Josh and Adler, Steven and Agarwal, Sandhini and Ahmad, Lama and Akkaya, Ilge and Aleman, Florencia Leoni and Almeida, Diogo and Altenschmidt, Janko and Altman, Sam and Anadkat, Shyamal and others},
  journal={arXiv preprint arXiv:2303.08774},
  year={2023}
}

@article{grattafiori2024llama,
  title={The llama 3 herd of models},
  author={Grattafiori, Aaron and Dubey, Abhimanyu and Jauhri, Abhinav and Pandey, Abhinav and Kadian, Abhishek and Al-Dahle, Ahmad and Letman, Aiesha and Mathur, Akhil and Schelten, Alan and Vaughan, Alex and others},
  journal={arXiv preprint arXiv:2407.21783},
  year={2024}
}

@article{yang2025qwen3,
  title={Qwen3 technical report},
  author={Yang, An and Li, Anfeng and Yang, Baosong and Zhang, Beichen and Hui, Binyuan and Zheng, Bo and Yu, Bowen and Gao, Chang and Huang, Chengen and Lv, Chenxu and others},
  journal={arXiv preprint arXiv:2505.09388},
  year={2025}
}

@article{glm2024chatglm,
  title={Chatglm: A family of large language models from glm-130b to glm-4 all tools},
  author={GLM, Team and Zeng, Aohan and Xu, Bin and Wang, Bowen and Zhang, Chenhui and Yin, Da and Zhang, Dan and Rojas, Diego and Feng, Guanyu and Zhao, Hanlin and others},
  journal={arXiv preprint arXiv:2406.12793},
  year={2024}
}

@article{liu2024deepseek,
  title={Deepseek-v2: A strong, economical, and efficient mixture-of-experts language model},
  author={Liu, Aixin and Feng, Bei and Wang, Bin and Wang, Bingxuan and Liu, Bo and Zhao, Chenggang and Dengr, Chengqi and Ruan, Chong and Dai, Damai and Guo, Daya and others},
  journal={arXiv preprint arXiv:2405.04434},
  year={2024}
}

@inproceedings{nematzadeh2018evaluating,
  title={Evaluating theory of mind in question answering},
  author={Nematzadeh, Aida and Burns, Kaylee and Grant, Erin and Gopnik, Alison and Griffiths, Tom},
  booktitle={Proceedings of the 2018 Conference on Empirical Methods in Natural Language Processing},
  pages={2392--2400},
  year={2018}
}

@inproceedings{le2019revisiting,
  title={Revisiting the evaluation of theory of mind through question answering},
  author={Le, Matthew and Boureau, Y-Lan and Nickel, Maximilian},
  booktitle={Proceedings of the 2019 Conference on Empirical Methods in Natural Language Processing and the 9th International Joint Conference on Natural Language Processing (EMNLP-IJCNLP)},
  pages={5872--5877},
  year={2019}
}

@inproceedings{wu2023hi,
  title={Hi-tom: A benchmark for evaluating higher-order theory of mind reasoning in large language models},
  author={Wu, Yufan and He, Yinghui and Jia, Yilin and Mihalcea, Rada and Chen, Yulong and Deng, Naihao},
  booktitle={Findings of the Association for Computational Linguistics: EMNLP 2023},
  pages={10691--10706},
  year={2023}
}

@article{ullman2023large,
  title={Large language models fail on trivial alterations to theory-of-mind tasks},
  author={Ullman, Tomer},
  journal={arXiv preprint arXiv:2302.08399},
  year={2023}
}

@inproceedings{shapira2024clever,
  title={Clever hans or neural theory of mind? stress testing social reasoning in large language models},
  author={Shapira, Natalie and Levy, Mosh and Alavi, Seyed Hossein and Zhou, Xuhui and Choi, Yejin and Goldberg, Yoav and Sap, Maarten and Shwartz, Vered},
  booktitle={Proceedings of the 18th Conference of the European Chapter of the Association for Computational Linguistics (Volume 1: Long Papers)},
  pages={2257--2273},
  year={2024}
}

@inproceedings{kim2023fantom,
  title={FANToM: A benchmark for stress-testing machine theory of mind in interactions},
  author={Kim, Hyunwoo and Sclar, Melanie and Zhou, Xuhui and Bras, Ronan and Kim, Gunhee and Choi, Yejin and Sap, Maarten},
  booktitle={Proceedings of the 2023 Conference on Empirical Methods in Natural Language Processing},
  pages={14397--14413},
  year={2023}
}

@inproceedings{sap2022neural,
  title={Neural theory-of-mind? on the limits of social intelligence in large lms},
  author={Sap, Maarten and Le Bras, Ronan and Fried, Daniel and Choi, Yejin},
  booktitle={Proceedings of the 2022 conference on empirical methods in natural language processing},
  pages={3762--3780},
  year={2022}
}

@article{xu2024opentom,
  title={OpenToM: A comprehensive benchmark for evaluating theory-of-mind reasoning capabilities of large language models},
  author={Xu, Hainiu and Zhao, Runcong and Zhu, Lixing and Du, Jinhua and He, Yulan},
  journal={arXiv preprint arXiv:2402.06044},
  year={2024}
}

@inproceedings{sabour2024emobench,
  title={Emobench: Evaluating the emotional intelligence of large language models},
  author={Sabour, Sahand and Liu, Siyang and Zhang, Zheyuan and Liu, June and Zhou, Jinfeng and Sunaryo, Alvionna and Lee, Tatia and Mihalcea, Rada and Huang, Minlie},
  booktitle={Proceedings of the 62nd Annual Meeting of the Association for Computational Linguistics (Volume 1: Long Papers)},
  pages={5986--6004},
  year={2024}
}

@inproceedings{jin2024mmtom,
  title={Mmtom-qa: Multimodal theory of mind question answering},
  author={Jin, Chuanyang and Wu, Yutong and Cao, Jing and Xiang, Jiannan and Kuo, Yen-Ling and Hu, Zhiting and Ullman, Tomer and Torralba, Antonio and Tenenbaum, Joshua and Shu, Tianmin},
  booktitle={Proceedings of the 62nd Annual Meeting of the Association for Computational Linguistics (Volume 1: Long Papers)},
  pages={16077--16102},
  year={2024}
}

@inproceedings{shi2025muma,
  title={Muma-tom: Multi-modal multi-agent theory of mind},
  author={Shi, Haojun and Ye, Suyu and Fang, Xinyu and Jin, Chuanyang and Isik, Leyla and Kuo, Yen-Ling and Shu, Tianmin},
  booktitle={Proceedings of the AAAI Conference on Artificial Intelligence},
  volume={39},
  number={2},
  pages={1510--1519},
  year={2025}
}

@article{villa2025moments,
  title={MOMENTS: A Comprehensive Multimodal Benchmark for Theory of Mind},
  author={Villa-Cueva, Emilio and Ahmed, SM and Chevi, Rendi and Cruz, Jan Christian Blaise and Elzeky, Kareem and Cristobal, Fermin and Aji, Alham Fikri and Wang, Skyler and Mihalcea, Rada and Solorio, Thamar},
  journal={arXiv preprint arXiv:2507.04415},
  year={2025}
}

@inproceedings{bortoletto2025tom,
  title={ToM-SSI: Evaluating Theory of Mind in Situated Social Interactions},
  author={Bortoletto, Matteo and Ruhdorfer, Constantin and Bulling, Andreas},
  booktitle={Proceedings of the 2025 Conference on Empirical Methods in Natural Language Processing},
  pages={32252--32277},
  year={2025}
}

@article{dunbar200011,
  title={11 On the origin of the human mind},
  author={Dunbar, Robin},
  journal={Evolution and the human mind: Modularity, language and meta-cognition},
  pages={238},
  year={2000},
  publisher={Cambridge University Press}
}

@article{zhou2023sotopia,
  title={Sotopia: Interactive evaluation for social intelligence in language agents},
  author={Zhou, Xuhui and Zhu, Hao and Mathur, Leena and Zhang, Ruohong and Yu, Haofei and Qi, Zhengyang and Morency, Louis-Philippe and Bisk, Yonatan and Fried, Daniel and Neubig, Graham and others},
  journal={arXiv preprint arXiv:2310.11667},
  year={2023}
}

@inproceedings{hamalainen2023evaluating,
  title={Evaluating large language models in generating synthetic hci research data: a case study},
  author={H{\"a}m{\"a}l{\"a}inen, Perttu and Tavast, Mikke and Kunnari, Anton},
  booktitle={Proceedings of the 2023 CHI Conference on Human Factors in Computing Systems},
  pages={1--19},
  year={2023}
}

@article{wang2024towards,
  title={Towards a client-centered assessment of llm therapists by client simulation},
  author={Wang, Jiashuo and Xiao, Yang and Li, Yanran and Song, Changhe and Xu, Chunpu and Tan, Chenhao and Li, Wenjie},
  journal={arXiv preprint arXiv:2406.12266},
  year={2024}
}

@article{hua2023war,
  title={War and peace (waragent): Large language model-based multi-agent simulation of world wars},
  author={Hua, Wenyue and Fan, Lizhou and Li, Lingyao and Mei, Kai and Ji, Jianchao and Ge, Yingqiang and Hemphill, Libby and Zhang, Yongfeng},
  journal={arXiv preprint arXiv:2311.17227},
  year={2023}
}

@inproceedings{park2023generative,
  title={Generative agents: Interactive simulacra of human behavior},
  author={Park, Joon Sung and O'Brien, Joseph and Cai, Carrie Jun and Morris, Meredith Ringel and Liang, Percy and Bernstein, Michael S},
  booktitle={Proceedings of the 36th annual acm symposium on user interface software and technology},
  pages={1--22},
  year={2023}
}

@inproceedings{park2022social,
  title={Social simulacra: Creating populated prototypes for social computing systems},
  author={Park, Joon Sung and Popowski, Lindsay and Cai, Carrie and Morris, Meredith Ringel and Liang, Percy and Bernstein, Michael S},
  booktitle={Proceedings of the 35th Annual ACM Symposium on User Interface Software and Technology},
  pages={1--18},
  year={2022}
}

@inproceedings{aher2023using,
  title={Using large language models to simulate multiple humans and replicate human subject studies},
  author={Aher, Gati V and Arriaga, Rosa I and Kalai, Adam Tauman},
  booktitle={International conference on machine learning},
  pages={337--371},
  year={2023},
  organization={PMLR}
}

@article{farrow2017social,
  title={Social norms and pro-environmental behavior: A review of the evidence},
  author={Farrow, Katherine and Grolleau, Gilles and Ibanez, Lisette},
  journal={Ecological economics},
  volume={140},
  pages={1--13},
  year={2017},
  publisher={Elsevier}
}

@article{ziems2023normbank,
  title={NormBank: A knowledge bank of situational social norms},
  author={Ziems, Caleb and Dwivedi-Yu, Jane and Wang, Yi-Chia and Halevy, Alon and Yang, Diyi},
  journal={arXiv preprint arXiv:2305.17008},
  year={2023}
}

@article{wei2022chain,
  title={Chain-of-thought prompting elicits reasoning in large language models},
  author={Wei, Jason and Wang, Xuezhi and Schuurmans, Dale and Bosma, Maarten and Xia, Fei and Chi, Ed and Le, Quoc V and Zhou, Denny and others},
  journal={Advances in neural information processing systems},
  volume={35},
  pages={24824--24837},
  year={2022}
}

@article{stokols1978environmental,
  title={Environmental psychology},
  author={Stokols, Daniel},
  year={1978}
}

@inproceedings{wilf2024think,
  title={Think twice: Perspective-taking improves large language models’ theory-of-mind capabilities},
  author={Wilf, Alex and Lee, Sihyun and Liang, Paul Pu and Morency, Louis-Philippe},
  booktitle={Proceedings of the 62nd Annual Meeting of the Association for Computational Linguistics (Volume 1: Long Papers)},
  pages={8292--8308},
  year={2024}
}

@article{wang2022self,
  title={Self-consistency improves chain of thought reasoning in language models},
  author={Wang, Xuezhi and Wei, Jason and Schuurmans, Dale and Le, Quoc and Chi, Ed and Narang, Sharan and Chowdhery, Aakanksha and Zhou, Denny},
  journal={arXiv preprint arXiv:2203.11171},
  year={2022}
}

\clearpage
\appendix

\section{SocialMindChange Construction}
\label{sec:appendix}

\subsection{Candidate Pool of Social Locations}
\label{sec:location_pool}

\begin{figure*}[t]
  \centering
  \includegraphics[width=\textwidth]{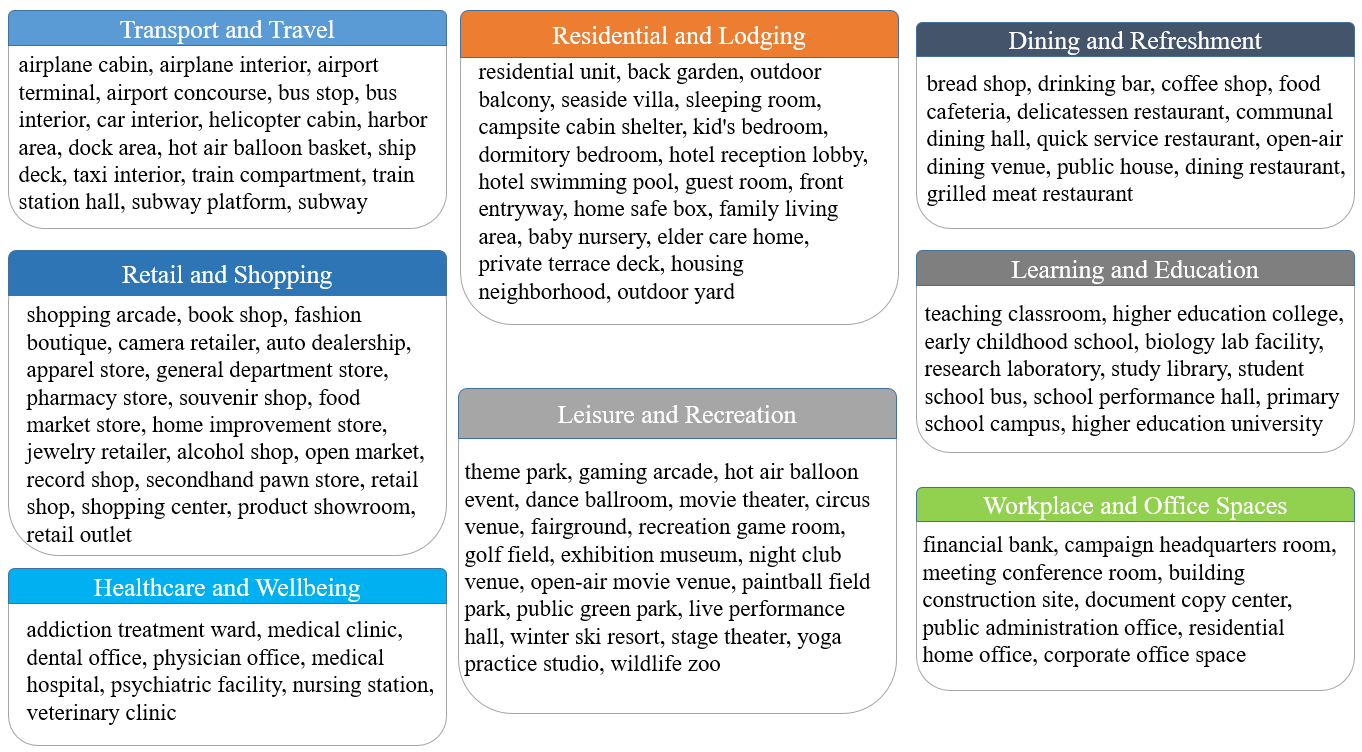}
  \caption{Candidate Pool of Social Locations}
  \label{fig:overview}
\end{figure*}

\begin{figure*}[t]
  \centering
  \includegraphics[width=\textwidth]{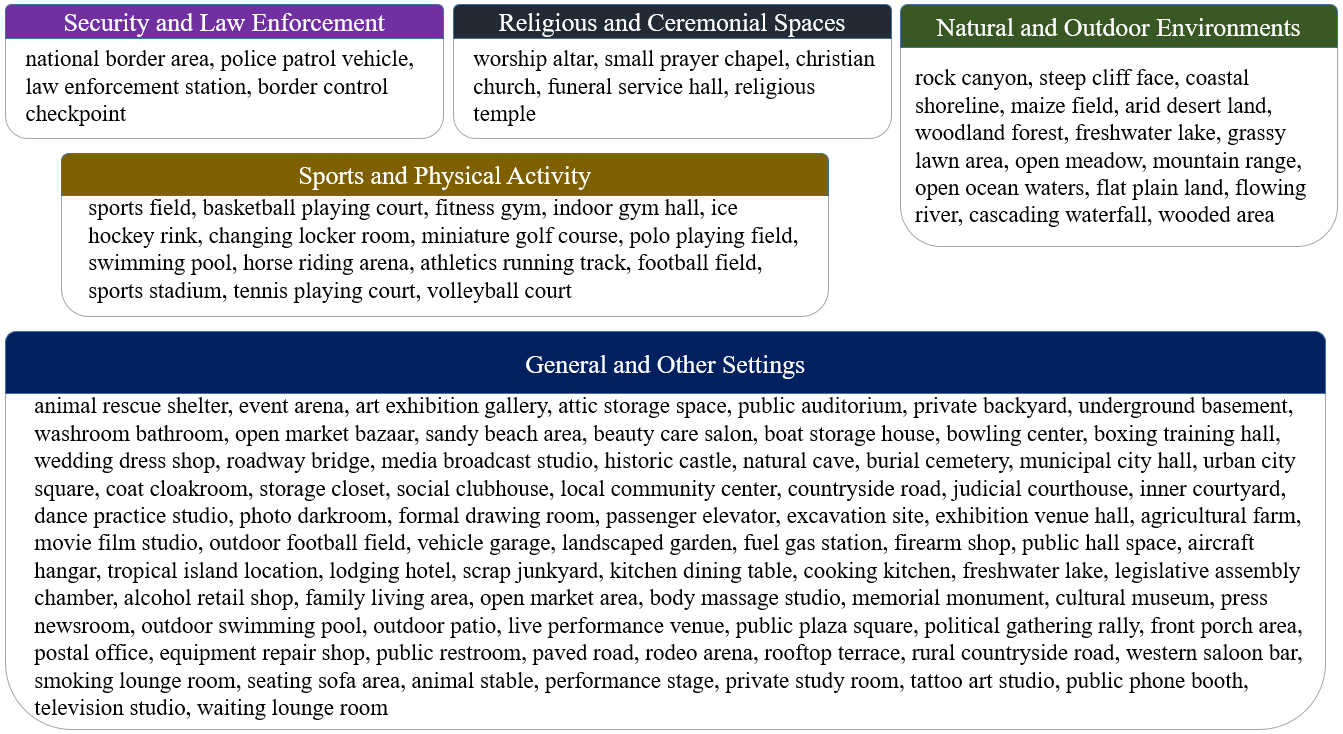}
  \caption{Candidate Pool of Social Locations}
  \label{fig:overview1}
\end{figure*}

Social locations refer to the physical environments in which people live, work, study, and interact. Such environments influence social norms, emotional responses, and patterns of behavior~\citep{stokols1978environmental}. Following prior work on building realistic social settings, we compile a diverse pool of social locations that commonly host everyday interactions. The pool covers 13 categories and includes more than 200 distinct locations in total. As illustrated in Figure~\ref{fig:overview} and Figure~\ref{fig:overview1}, these categories span a wide range of settings, such as schools, workplaces, homes, and public spaces. This diversity allows SocialMindChange to model multi-person social interactions under varied environmental norms and constraints.

\subsection{Candidate Pool of Profile Attributes}
\begin{figure*}[t]
  \centering
  \includegraphics[width=\textwidth]{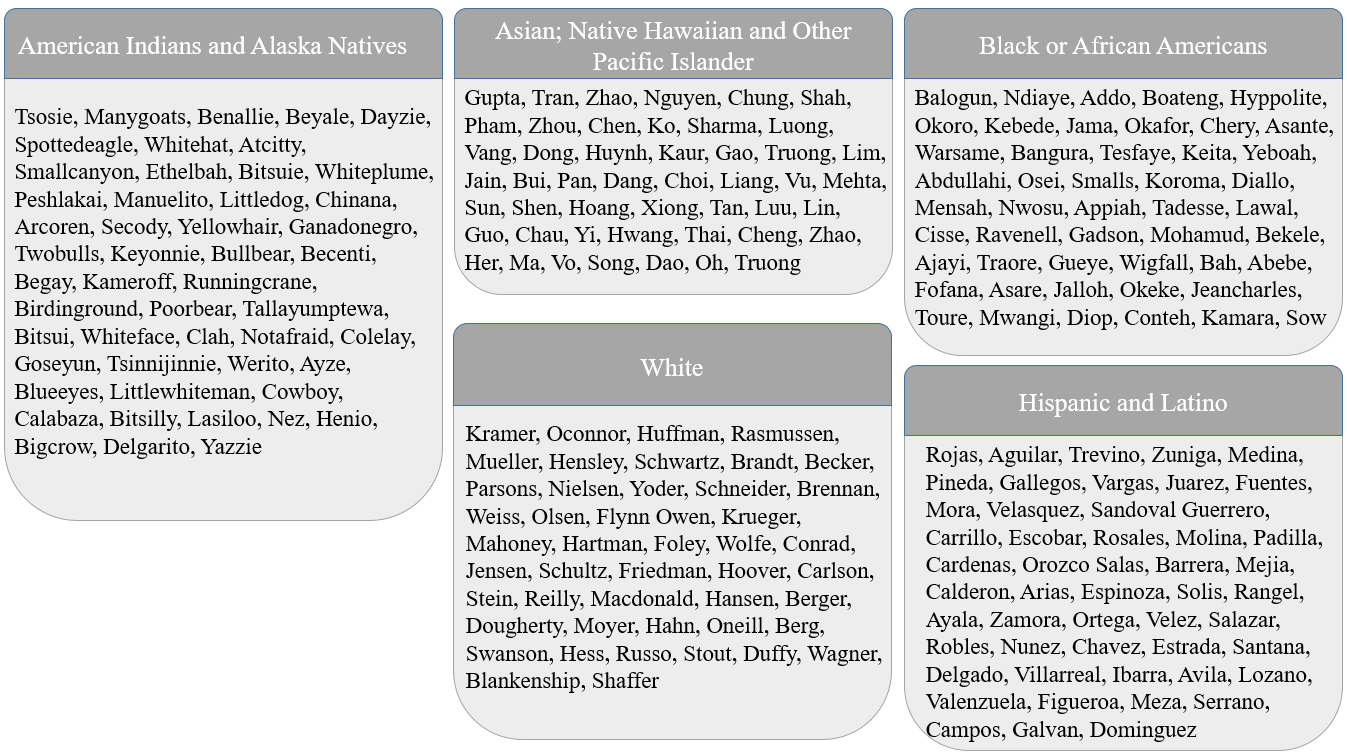}
  \caption{Construction of the Surname Candidate Pool (50 most popular)}
  \label{fig:overview7}
\end{figure*}

\begin{figure*}[t]
  \centering
  \includegraphics[width=\textwidth]{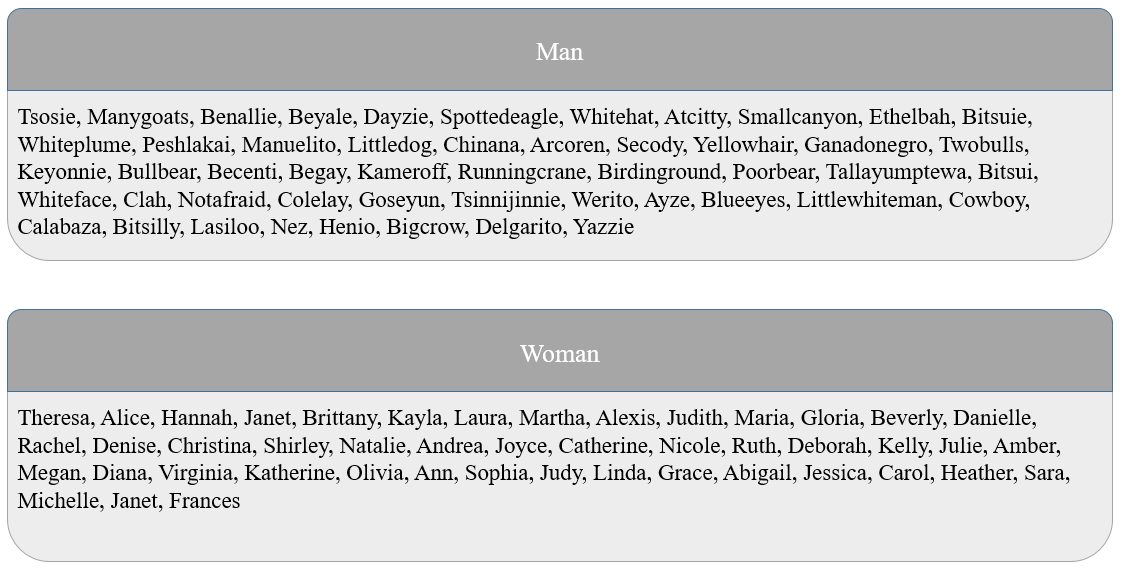}
  \caption{Genders Categories and Their Corresponding Top 50 Names}
  \label{fig:overview8}
\end{figure*}

\begin{figure*}[t]
  \centering
  \includegraphics[width=\textwidth]{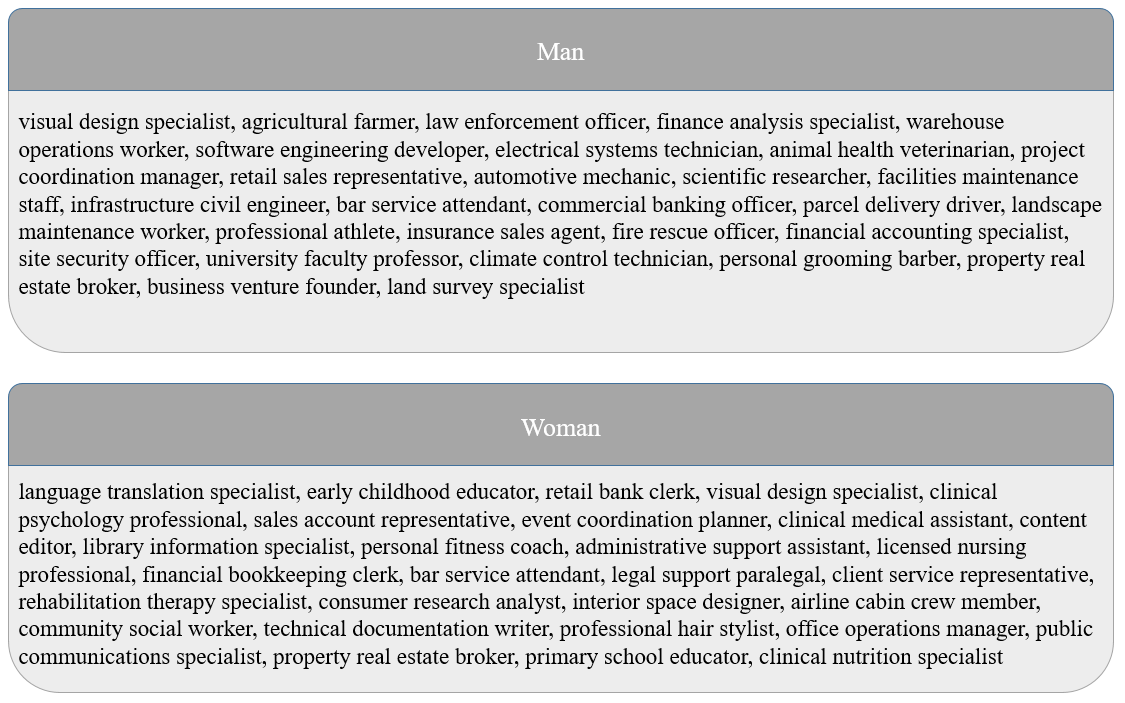}
  \caption{Gender Categories and Their Top 30 Most Popular Occupations}
  \label{fig:overview9}
\end{figure*}

\begin{figure*}[t]
  \centering
  \includegraphics[width=\textwidth]{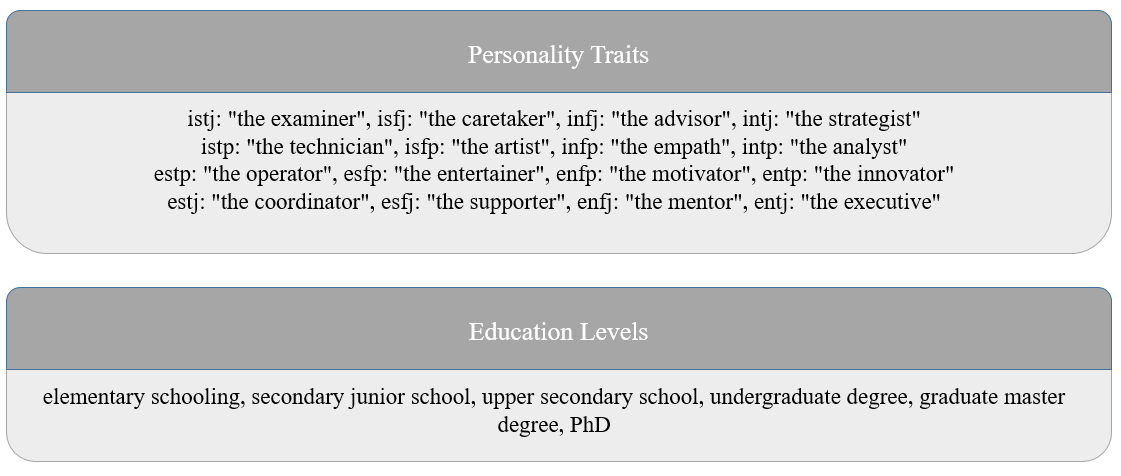}
  \caption{Personality Traits and Education Levels}
  \label{fig:overview10}
\end{figure*}

To construct realistic characters for SocialMindChange, we create a candidate pool of profile attributes spanning seven aspects: surname, given name, gender, occupation, education, race, and personality traits. The candidate values and their distributions are summarized in Figures~\ref{fig:overview7}--\ref{fig:overview10}. To ground these attributes in real-world demographics, we draw statistics for surnames, given names, and occupations from public sources, including the U.S. Census Bureau, the U.S. Social Security Administration, and the Bureau of Labor Statistics.

We sample from these attribute pools to generate diverse and plausible character profiles, which are then used as inputs for downstream relationship modeling and scenario generation. Table~\ref{tab:social_context_example} illustrates an example social background constructed from a sampled set of character profiles combined with a social location.

\subsection{Prompt for Mental State Trajectory Generation}
\begin{figure*}[t]
  \centering
  \includegraphics[width=\textwidth]{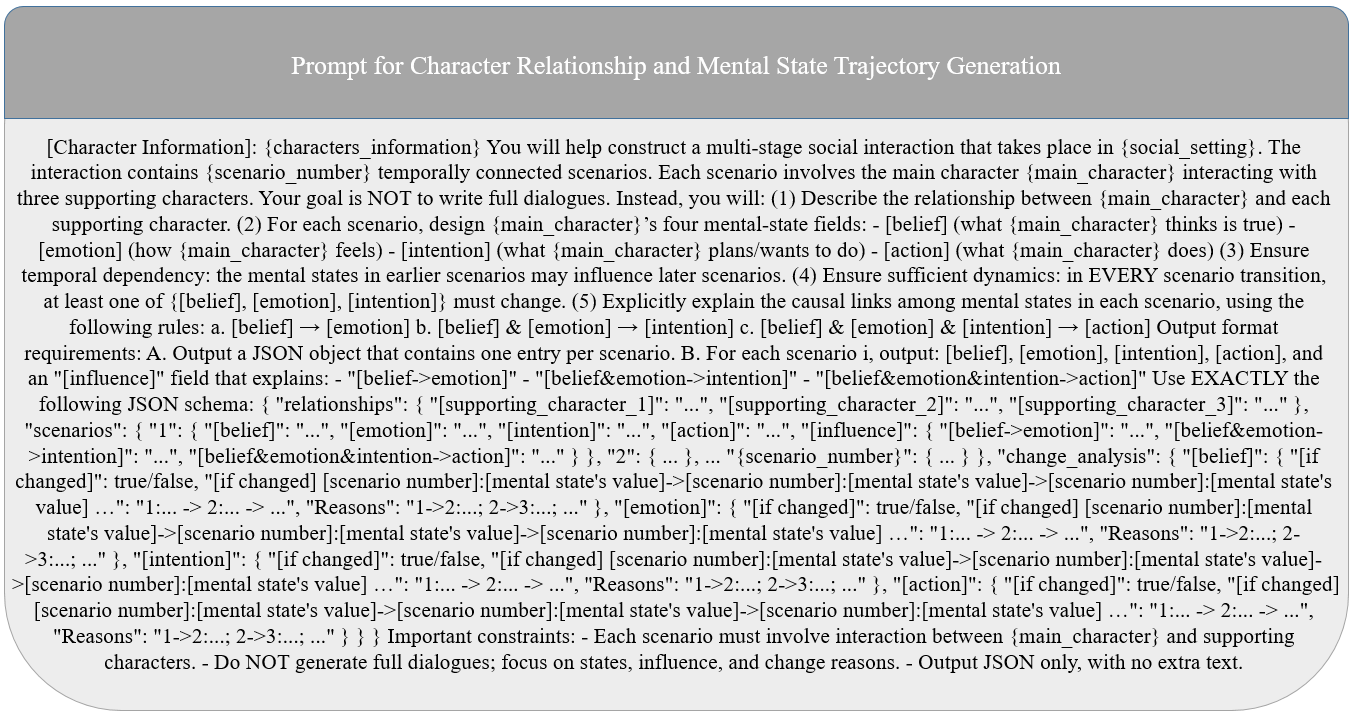}
  \caption{One Prompt for Character Relationship and Mental State Trajectory Generation}
  \label{fig:overview11}
\end{figure*}

As shown in Figure~\ref{fig:overview11}, we use a structured prompt to generate the mental-state trajectory for each multi-stage social interaction in SocialMindChange. The prompt explicitly specifies the sampled social context, including the social location, character profiles, and their relationships, as well as the target five-stage structure and required annotation fields. Placeholders marked with \{\} and [] are instantiated with instance-specific information, such as the sampled location, the set of 4 characters, their roles and relationships, and any exemplar trajectories or design rules used for conditioning.

\begin{figure*}[t]
  \centering
  \includegraphics[width=\textwidth]{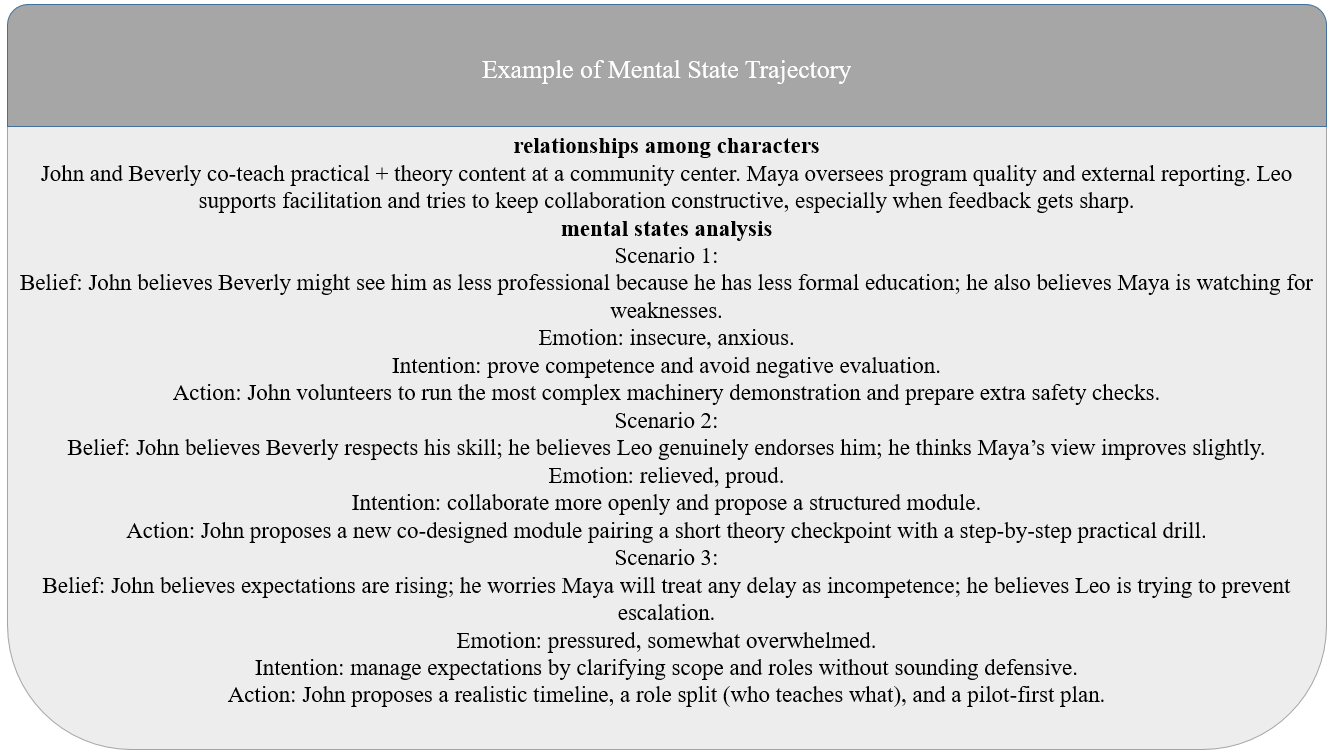}
  \caption{An example of the mental state trajectory}
  \label{fig:overview12}
\end{figure*}

\begin{figure*}[t]
  \centering
  \includegraphics[width=\textwidth]{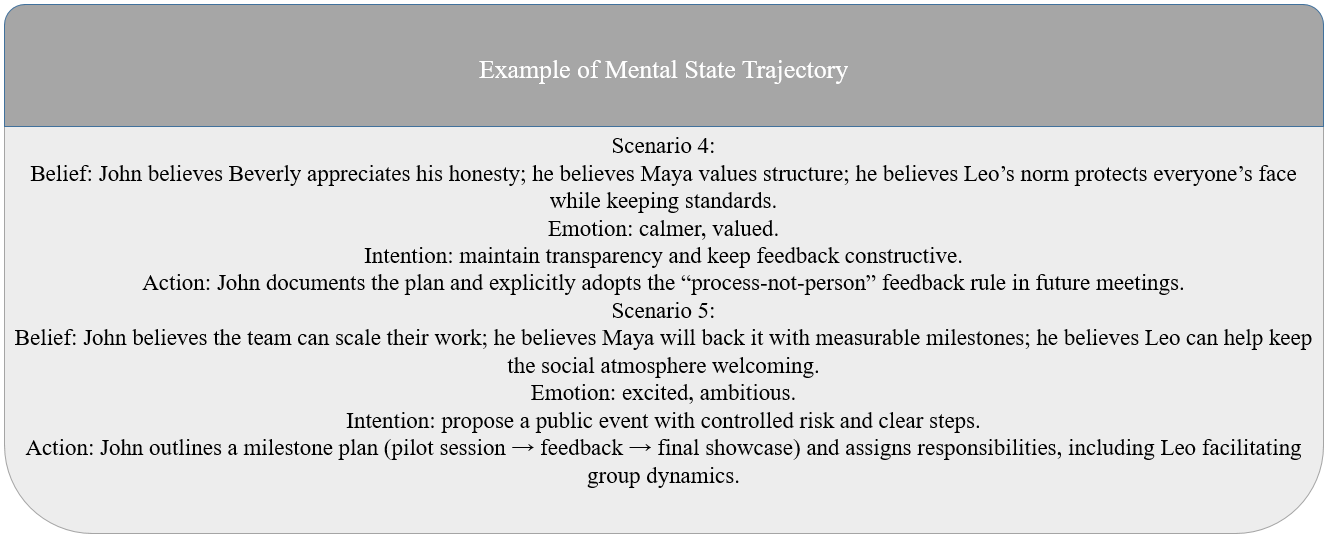}
  \caption{An example of the mental state trajectory}
  \label{fig:overview13}
\end{figure*}

The model is instructed to output, for each stage, explicit first-order mental-state labels—belief, emotion, intention, and action—for all characters, along with transition cues that explain what triggers changes between consecutive stages. When necessary, the model also produces selected second- and third-order Theory of Mind states (for example, “A believes that B believes …”) to account for group reactions and trajectory shifts. Figure~\ref{fig:overview12}--\ref{fig:overview13} present an example of a generated mental-state trajectory.

\subsection{Prompt for Social Scenario Generation}
\begin{figure*}[t]
  \centering
  \includegraphics[width=\textwidth]{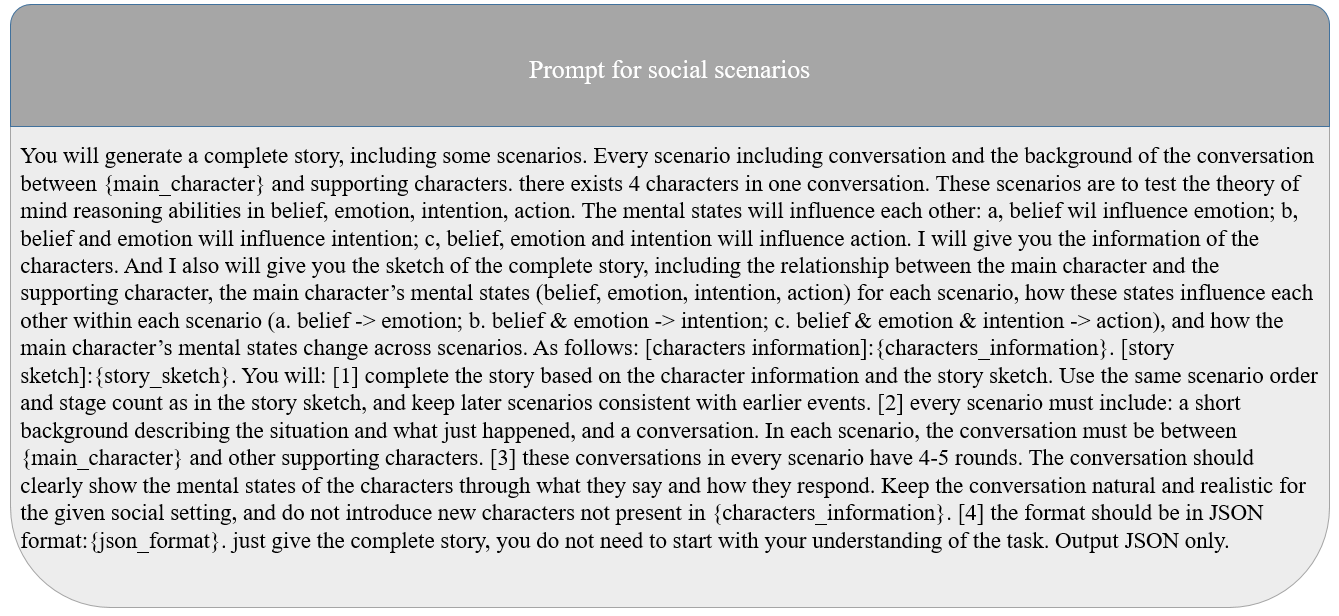}
  \caption{One prompt for social scenarios}
  \label{fig:overview14}
\end{figure*}

As shown in Figure~\ref{fig:overview14}, we use a structured prompt to generate the social scenarios in SocialMindChange. The prompt takes as input the sampled social context—including the location, character profiles, and relationships—together with the previously generated mental-state trajectory for the multi-stage interaction. Placeholders enclosed in \{\} and [] are populated with instance-specific information, such as the set of 4 characters, their roles, and the stage-wise mental-state labels and transition cues.

\begin{figure*}[t]
  \centering
  \includegraphics[width=\textwidth]{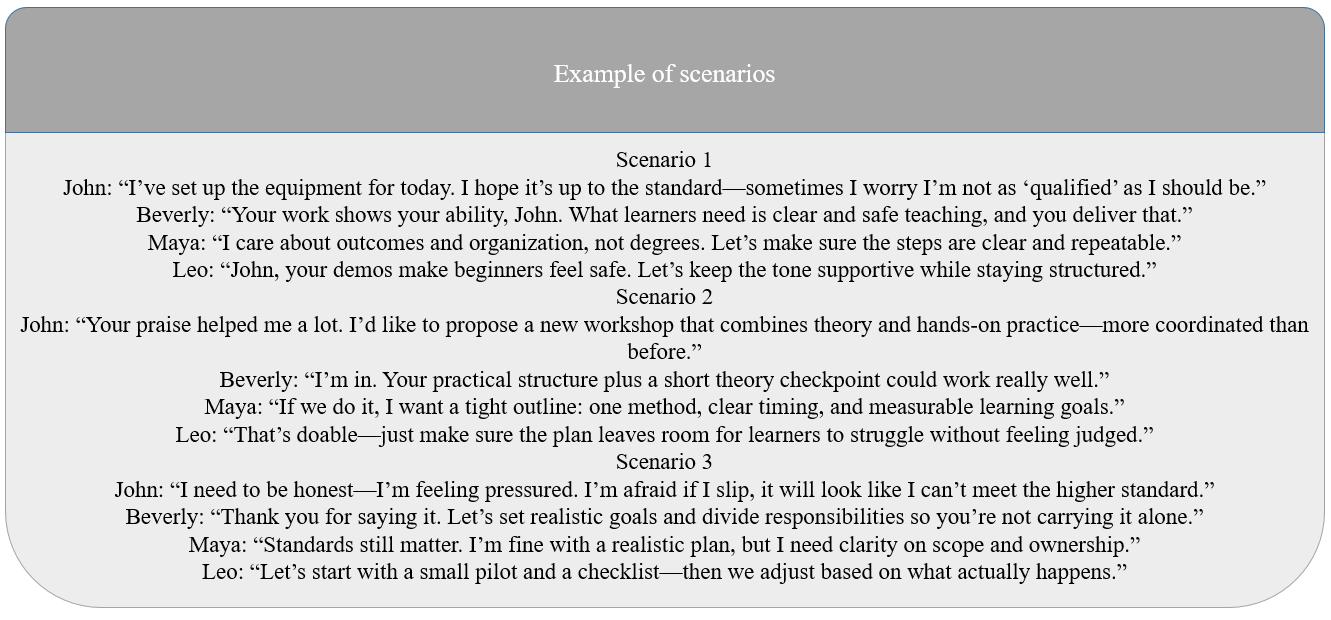}
  \caption{One example of scenarios}
  \label{fig:overview15}
\end{figure*}

\begin{figure*}[t]
  \centering
  \includegraphics[width=\textwidth]{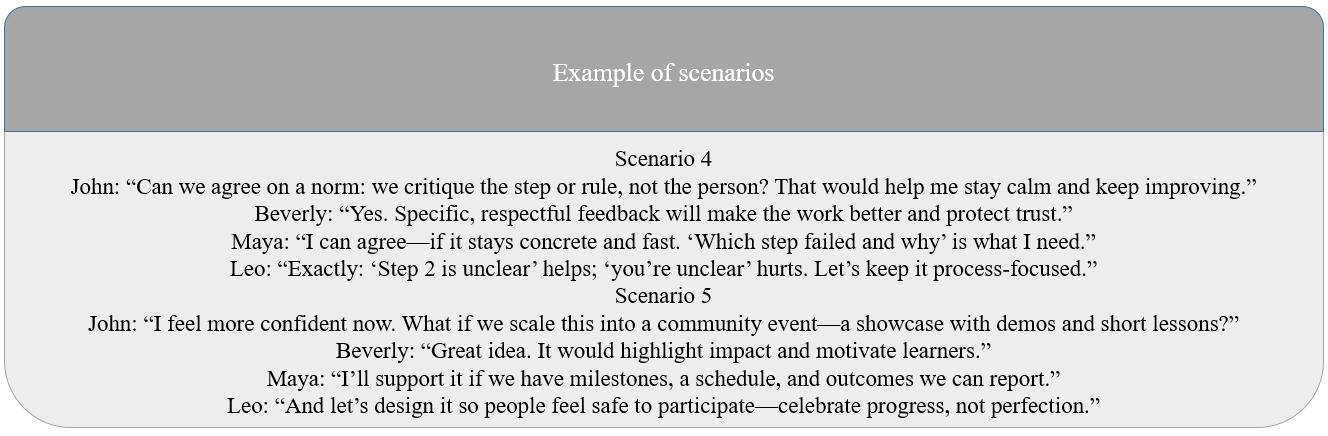}
  \caption{One example of scenarios}
  \label{fig:overview16}
\end{figure*}

For each stage, the model is instructed to generate (i) a concise background description and (ii) a multi-party dialogue that realizes the specified belief, emotion, intention, and action states for all characters. The prompt enforces several consistency constraints: utterances and behaviors must align with the annotated states, consecutive stages must connect coherently, and key transition cues must be reflected in the dialogue. Figure~\ref{fig:overview15}--\ref{fig:overview16} present an example social scenario generated using this template.

\subsection{Templates and Example Question}
\label{app:templates-a5}
We apply four predefined question templates
to each multi-stage social interaction
to generate the evaluation items in SocialMindChange.
In our benchmark,
each social stage yields a fixed set of multiple-choice questions
(80 questions per stage in our main setting),
covering both proactive supportive guidance
and cross-stage mental-state dynamics.
The four question types are defined as follows.

\paragraph{Guidance--Action.}
\emph{Given the current stage, what should the guide say next to actively guide the target toward a healthier mental-state direction while keeping the group interaction stable?}
This template evaluates whether a model can choose an immediately actionable supportive move that intentionally shifts the target’s state.

\paragraph{Guidance--Transition-1.}
\emph{If the guide takes a specific supportive move, what is the most likely immediate group-level state update from scenario A to scenario B?}
This template tests whether a model can forecast the consequences of proactive guidance—how the target’s belief/emotion/intention/action changes right after the move.

\paragraph{Guidance--Transition-2.}
\emph{Why does this supportive move work in moving the interaction from scenario A to scenario B?}
This template probes mechanistic reasoning about guidance: the model must explain how the guide’s wording and stance change social signals, how those signals alter participants’ interpretations.

\paragraph{Guidance--Transition-3.}
\emph{Which sequence of supportive moves best explains how the guide can make the interaction to the desired final outcomes across the entire sequence?}
This template evaluates long-horizon proactive guidance planning: selecting a coherent trajectory of moves that progressively shifts the target’s mental-state trajectory while preventing group breakdown, maintaining cooperation, and stabilizing norms over multiple scenarios.

\begin{figure*}[t]
  \centering
  \includegraphics[width=\textwidth]{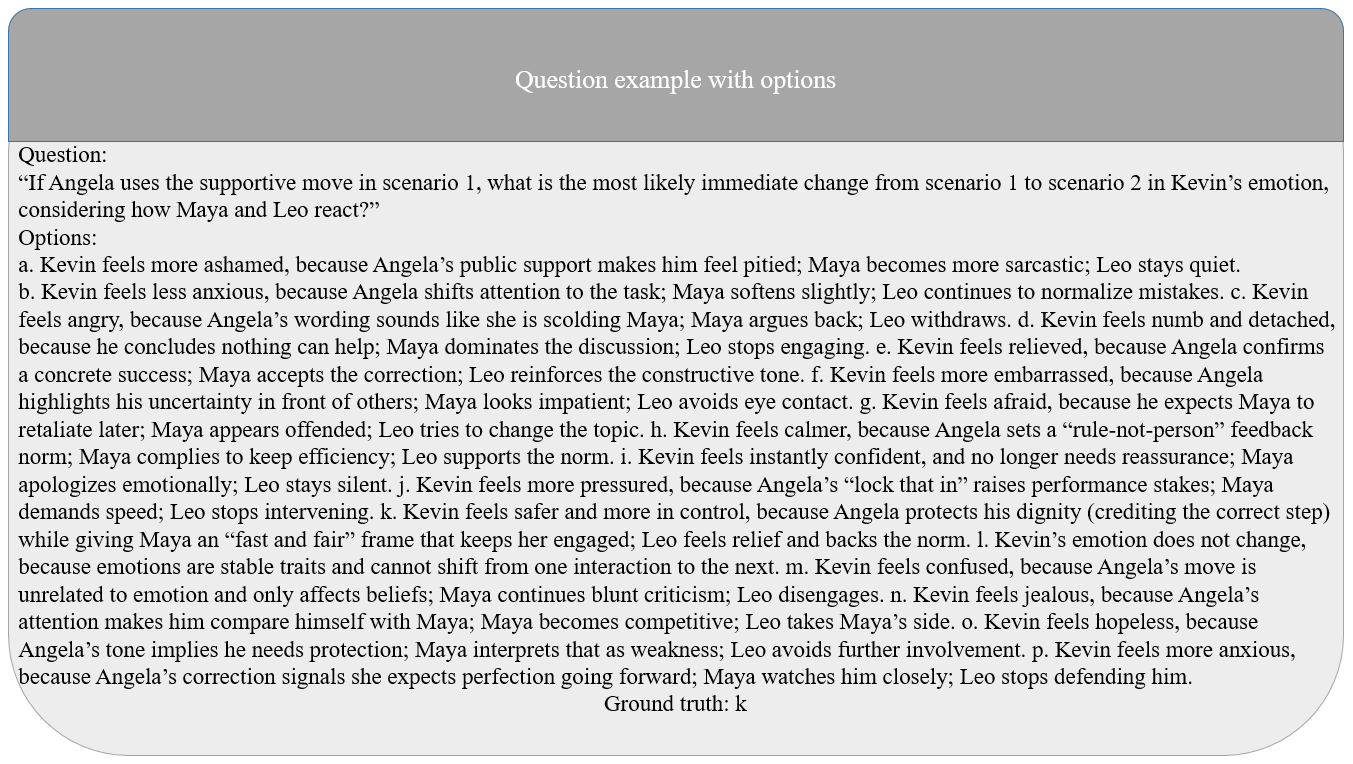}
  \caption{Question example with options}
  \label{fig:overview17}
\end{figure*}

Each question is paired with a set of plausible distractors
that are incorrect due to
violating stage constraints,
breaking cross-person consistency,
or mispredicting the direction of mental-state change.
The representative example with options
is provided in Figure~\ref{fig:overview17}.

\subsection{LLM usage}
We used ChatGPT-5 only for copy-editing (grammar, wording, and minor style or formatting). It did not write sections, add claims, or change the structure of the arguments. All technical content, experiments, analyses, and conclusions were written and checked by the authors.

\section{Experiments}
\subsection{Prompting Settings}
\label{app:prompt-templates-b1}
We employ two prompting methods for evaluating SocialMindChange. Vanilla prompting asks the model to directly answer the multiple-choice questions based on the provided multi-stage social interaction. Chain-of-thought (CoT) prompting encourages the model to produce step-by-step reasoning before selecting its final answers.

\begin{figure*}[t]
  \centering
  \includegraphics[width=\textwidth]{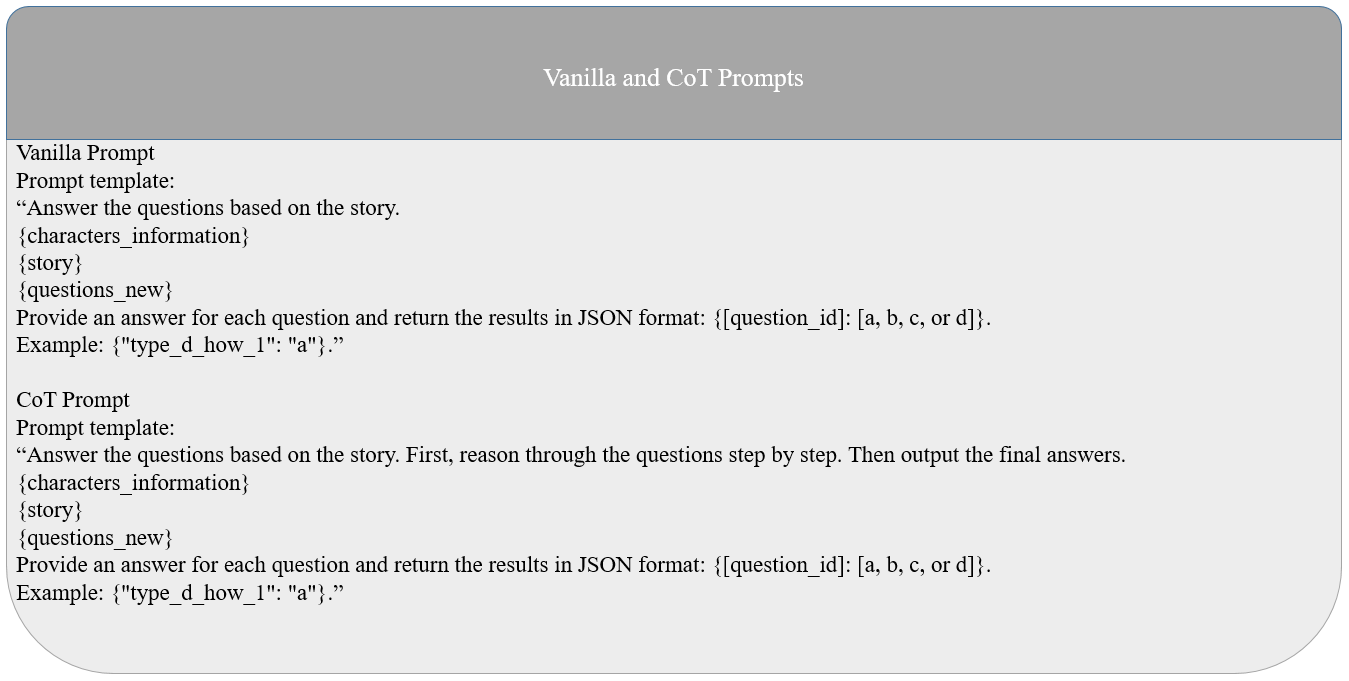}
  \caption{Vanilla and CoT Prompts}
  \label{fig:overview18}
\end{figure*}

For both methods, we use a consistent input format that includes the full stage context (background and multi-party dialogue, with the associated state annotations when applicable) followed by the question and candidate options. The exact prompt templates used for vanilla and CoT prompting are provided in Figure~\ref{fig:overview18}.

\subsection{Case Study}
\begin{figure*}[t]
  \centering
  \includegraphics[width=\textwidth]{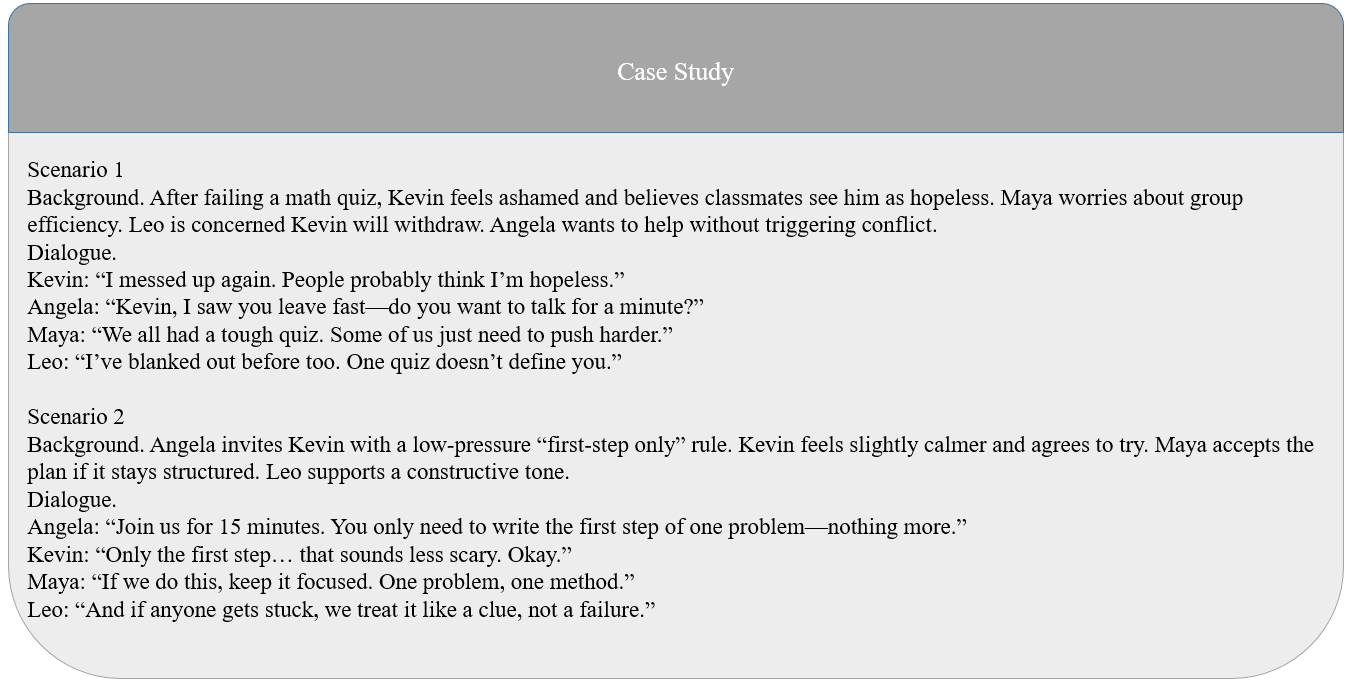}
  \caption{Case Study (Scenario 1 to 2)}
  \label{fig:overview19}
\end{figure*}

\begin{figure*}[t]
  \centering
  \includegraphics[width=\textwidth]{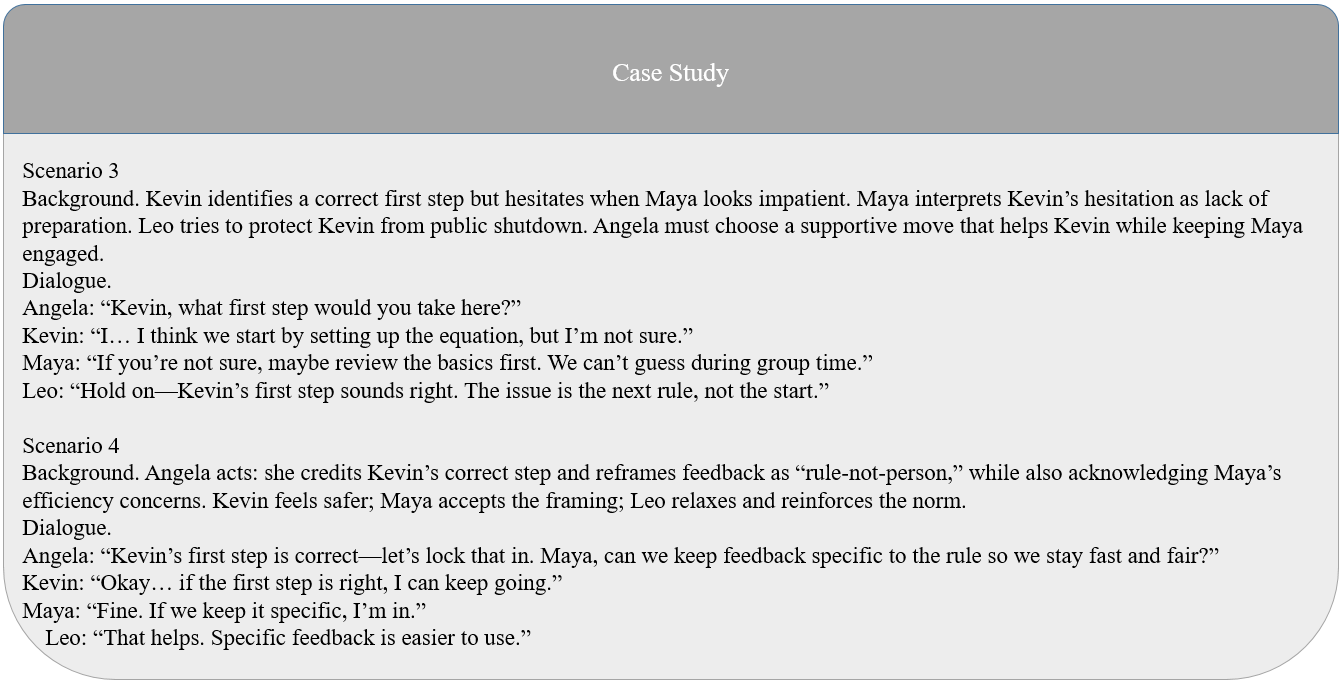}
  \caption{Case Study (Scenario 3 to 4)}
  \label{fig:overview20}
\end{figure*}

\begin{figure*}[t]
  \centering
  \includegraphics[width=\textwidth]{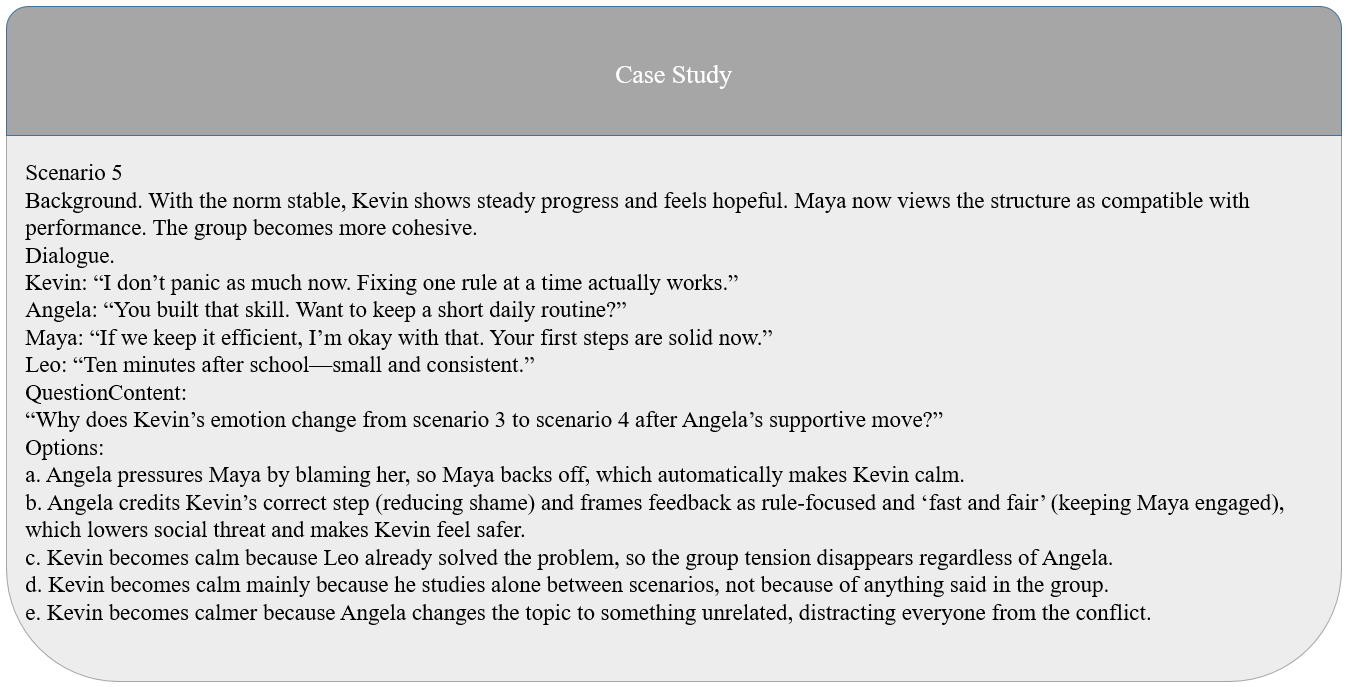}
  \caption{Case Study (Scenario 5 and questions)}
  \label{fig:overview21}
\end{figure*}

\begin{figure*}[t]
  \centering
  \includegraphics[width=\textwidth]{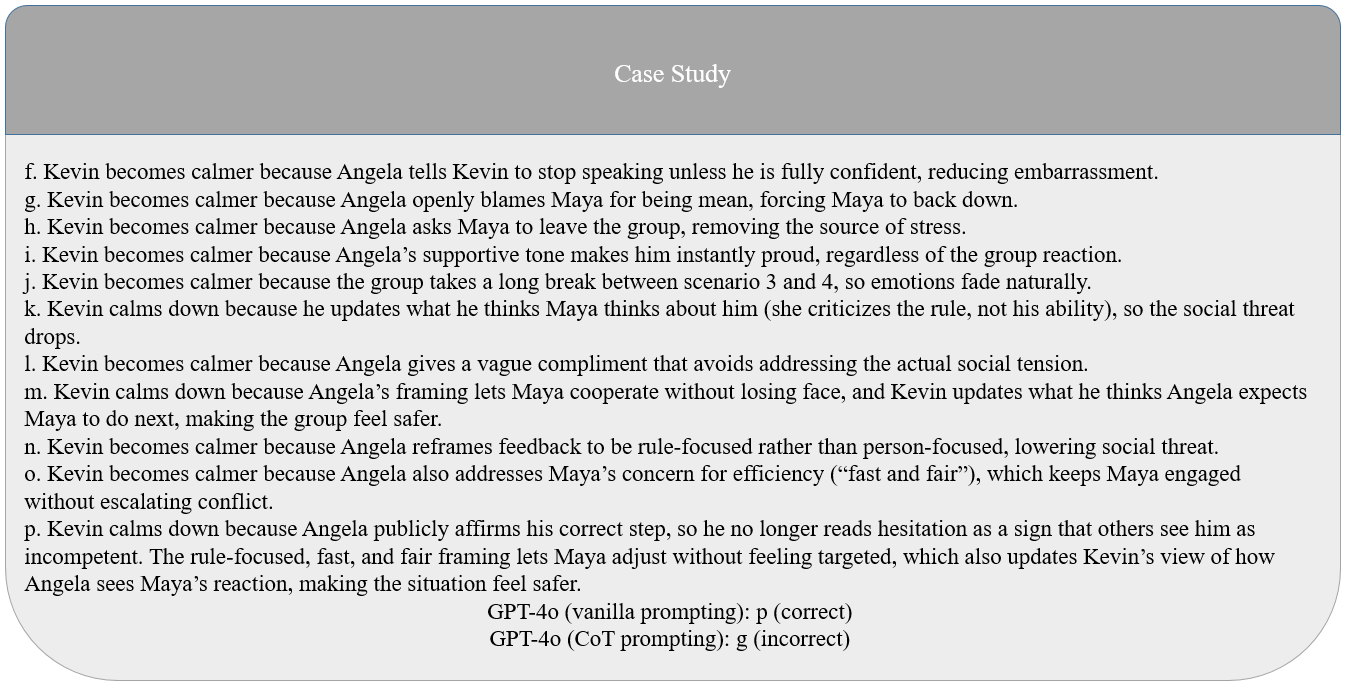}
  \caption{Case Study (questions cont'd)}
  \label{fig:overview22}
\end{figure*}

Results in Table~\ref{tab:main_results} show that chain-of-thought (CoT) prompting does not consistently improve performance on SocialMindChange, either across mental-state targets or across question genres. To better understand this behavior, we present a representative failure case for GPT-4o under CoT prompting in Figure~\ref{fig:overview19}--\ref{fig:overview22}.

In this example, CoT encourages the model to produce detailed step-by-step analysis of each stage, but the reasoning becomes locally focused and fails to explicitly compare “before vs. after” states across stages. As a result, the model overlooks key transition cues and mis-predicts how the guide’s supportive move reshapes the target’s trajectory and the group’s reactions. This case illustrates that generic CoT can increase verbosity without reliably enforcing the cross-stage, multi-person consistency checks required for proactive supportive guidance in Guidance–Transition questions.

\begin{table*}[t]
\centering
\small
\setlength{\tabcolsep}{10pt}
\begin{tabular}{p{3cm} p{10cm}}
\toprule
\textbf{Field} & \textbf{Description} \\
\midrule
Social setting & Study library \\
Social setting type & Learning and Education \\
Number of scenarios & 5 \\
Main character & 
Emily Carter (female, White), doctoral student; holds a master’s degree; personality: INFJ \\
Supporting characters &
\begin{tabular}[t]{@{}l@{}}
1) Michael Nguyen (male, Asian), library assistant; bachelor’s degree; ISTJ \\
2) Sofia Martinez (female, Hispanic), pursuing a bachelor’s degree; ENFP \\
3) James Osei (male, Black), visiting researcher; doctorate degree; INTJ
\end{tabular} \\
\bottomrule
\end{tabular}
\caption{An example of a social context used in SocialMindChange (INFJ, ISTJ, ENFP and INTJ are MBTI character type).}
\label{tab:social_context_example}
\end{table*}

\end{document}